%% file: neurips_2024.tex
\documentclass{article}

    \PassOptionsToPackage{numbers, compress}{natbib}
\usepackage[preprint]{neurips_2024}

\usepackage[utf8]{inputenc} % allow utf-8 input
\usepackage[T1]{fontenc}    % use 8-bit T1 fonts
\usepackage{hyperref}       % hyperlinks
\usepackage{url}            % simple URL typesetting
\usepackage{booktabs}       % professional-quality tables
\usepackage{amsfonts}       % blackboard math symbols
\usepackage{nicefrac}       % compact symbols for 1/2, etc.
\usepackage{microtype}      % microtypography
\usepackage{xcolor}         % colors

\usepackage{graphicx}
\usepackage{amsmath}
\usepackage{multirow}
\usepackage{booktabs}
\usepackage{subcaption}
\usepackage{listings}

\usepackage{graphicx, amsmath, amssymb, caption, subcaption, multirow, overpic, textpos}

\title{Multi-layer Learnable Attention Mask for Multimodal Tasks}
\author{
    \begin{minipage}[t]{0.45\textwidth}
        \centering
        Wayner Barrios \\
        Dartmouth College
    \end{minipage}
    \hfill
    \begin{minipage}[t]{0.45\textwidth}
        \centering
        SouYoung Jin \\
        Dartmouth College
    \end{minipage}
}
\begin{document}

\maketitle

\input{sections/0_abstract}    
\input{sections/1_intro}
\input{sections/2_related}
\input{sections/3_method}

\input{sections/4_experiment}
\input{sections/5_conclusion}

\bibliographystyle{plainnat}
\bibliography{main}
%\input{sections/checklist}

% % Appendix
% \newpage
 \input{sections/6_appendix}

% \input{sections/broader_impacts}
% \input{sections/limitation}
% \bibliographystyle{plainnat}
% \bibliography{main}
% \input{sections/checklist}

\end{document}

%% file: sections/0_abstract.tex
\begin{abstract}
While the Self-Attention mechanism in the Transformer model has  proven to be effective in many domains, we observe that it is less effective in more diverse settings (e.g. multimodality) due to the varying granularity of each token and the high computational demands of lengthy sequences. To address the challenges, we introduce the Learnable Attention Mask (LAM), strategically designed to globally regulate attention maps and prioritize critical tokens within the sequence. Leveraging the Self-Attention module in a BERT-like transformer network, our approach adeptly captures associations between tokens. The extension of the LAM to a multi-layer version accommodates the varied information aspects embedded at each layer of the Transformer network. Comprehensive experimental validation on various datasets, such as MADv2, QVHighlights, ImageNet 1K, and MSRVTT, demonstrates the efficacy of the LAM, exemplifying its ability to enhance model performance while mitigating redundant computations. This pioneering approach presents a significant advancement in enhancing the understanding of complex scenarios, such as in movie understanding.
% \keywords{Multi-modal learning \and Attention mechanisms}
\end{abstract}

%% file: sections/1_intro.tex
\section{Introduction}
\label{sec:intro}

The evolution of deep learning has empowered us to navigate increasingly complex scenarios, where many of which require digesting information from diverse sources, such as videos, images, audio, and text. One such scenario lies in understanding movie scenes~\cite{mad,autoad1,autoad2,Barrios_2023_ICCV,lsmdc,Xiao_2022_CVPR,Islam_2023_CVPR,Chen_2023_CVPR}, where models aim to extract meaningful insights from multiple modalities. 

% Consider a movie scene represented by video and audio tokens. While these tokens naturally align in time, each one can be associated with any other tokens presented in the scene. For example, in Figure~\ref{fig:LAM_teaser}(a), Mary and Tim engage in conversation, during which Tim requests Mary’s phone number. The scene is accompanied by cell phone typing sounds in the background. As they continue discussing their appearance, a background shout sound occurs. In the subsequent scene, we discover that it is indeed Mary’s friend, Joanna, who had called Mary in the previous scene. Thus, the significance of ``Joanna’s shouts’’ can be linked to the tokens in the subsequent scene. This type of attention can be captured by the Self-Attention module in the BERT-like transformer network~\cite{transformer,devlin2018bert}, wherein the model's input comprises a sequence of tokens from two different modalities.  

Consider a movie scene represented by video and audio tokens. While these tokens naturally align in time, each one can be associated with any other tokens presented in the scene, as in Figure~\ref{fig:LAM_teaser}(a).  
While the Self-Attention module is effective for computing local associations between tokens, we have observed several drawbacks in the current attention mechanism, especially when tokens originate from diverse modalities. Firstly, different modalities introduce varying granularities of information, leading to potential challenges. Each token in one modality may be associated with multiple tokens in the other modality. Such associations can extend beyond one-to-one correspondences, forming between sub-sequences of tokens in each modality. In Figure~\ref{fig:LAM_teaser}(a), ``Joanna’s shouts” might not be associated with a single video token but with several. Moreover, while longer sequence of tokens generally offers richer information, the computational demands of attention mechanisms increase with the input length of tokens. This constraint hinders the effective processing of a higher number of tokens. 
%Meanwhile, densely sampling consecutive frames may lead to redundant or even irrelevant information, potentially impacting performance. By mitigating redundant computations during training through the masking of duplicates, we can alleviate the limitations associated with longer inputs.

\input{figures/LAM_teaser.tex}
Our method stems from the empirical observation that not all tokens in complex input sequences carry equal importance. While prior works like ~\cite{mask_networks,MDMAN,adapt_mask,self_mask} have demonstrated the effectiveness of dynamically updated masking mechanisms, this concept remains relatively unexplored in the computer vision domain, with only a few studies such as ~\cite{swinbert} delving into it. This gap in vision research has motivated our comprehensive analysis of the impact of dynamic token masking across diverse vision tasks.

To tackle the challenges posed by complex input sequences, we propose the \textbf{\underline{L}earnable \underline{A}ttention \underline{M}ask (LAM)} – a novel mechanism that dynamically generates masks to regulate attention maps and prioritize critical tokens based on their contextual significance. By processing the entire input sequence, LAM enables efficient token inspection and dynamic prioritization tailored to each sequence. This adaptive masking technique seamlessly integrates into existing Transformer Encoder architectures, offering a flexible solution to enhance transformer-based models across diverse applications. Given the widespread adoption of transformers, LAM's plug-and-play nature could benefit researchers by enabling performance improvements through effortless integration.

The LAM method employs linear layers that take a complex token sequence as input, which can be either single-modal or multimodal. The output is a mask of size $L_t \times L_t$, where $L_t$ represents the length of the input sequence $T$. The generated mask can be applied globally across all transformer layers in the stack, or it can be scaled individually for each transformer layer. 

%To overcome these challenges, we introduce the Learnable Attention Mask (LAM), a novel mechanism designed to generate a mask that regulates attention maps and prioritizes focus on critical tokens within long input sequences. The LAM processes the entire input sequence, enabling our model to efficiently inspect all tokens and dynamically prioritize them according to their significance, even in prolonged sequences. This adaptive masking approach can be seamlessly integrated into existing Transformer Encoder architectures, offering a flexible solution that can be applied to enhance any state-of-the-art transformer-based model across diverse domains, benchmarks, and tasks. Given the recent widespread adoption of transformer architectures, the LAM could prove helpful for the research community working on these models.

% WE NEED SOME DETAILS HERE. here!!
Eventually, the resulting mask, as illustrated in Figure~\ref{fig:LAM_teaser}(b), captures attention structures globally. This generated mask is then element-wise multiplied with the attention scores, allowing the masking out or prioritization of specific tokens. Furthermore, our observation that each layer in the Transformer Network embeds different information aspects motivates us to install the LAM module per stage, leading to the extension of the LAM to a multi-layer LAM.

We validate the effectiveness of our approach across various experimental settings. Initially, we assess the capability of the multi-layer LAM in multimodal settings, presenting results from audio description generation experiments on the MADv2 dataset~\cite{mad,autoad1}. Additionally, we apply our approach to Moment Retrieval and Highlight Detection tasks on the QVHighlight dataset~\cite{moment-detr}, incorporating both text and video inputs. Furthermore, we demonstrate that the LAM could enhance the performance in single-modality settings, such as image classification task in ImageNet 1K~\cite{imagenet} and video captioning task in MSRVTT~\cite{msrvtt} where a single modality is considered as input to the model. While the performance gain in single-modality settings is not significant, we demonstrate that our multi-layer LAM can be adopted in various scenarios. Finally, we analyze how the generated mask effectively regulates attention maps. 
% Across all experiments, our proposed approach with the generated mask boosts model performance.

In summary, our contributions are \textbf{three-fold}:
\begin{enumerate}
\item We propose the Learnable Attention Mask (LAM), an innovative mechanism that dynamically identifies and prioritizes the most significant tokens within intricate input sequences. By generating masks that regulate attention maps, LAM ensures that the model's focus is directed towards the critical tokens, optimizing performance on complex sequence processing tasks. With the widespread adoption of transformer architectures across various domains, the LAM's modular design allows for seamless integration into existing transformer encoder frameworks. This plug-and-play capability presents a valuable opportunity for researchers to leverage LAM's token prioritization capabilities, potentially enhancing model performance with minimal effort and architectural modifications.
\item Through extensive experiments across various benchmarks, including MAD, ImageNet-1K, MSRVTT, and QVHighlights, we empirically demonstrate the effectiveness of our LAM method, particularly when employed with multimodal encoders.
\item We analyze the output of the LAM and its influence on attention weight distributions, supplemented by qualitative analysis to provide insights into its behavior.
\end{enumerate}

%% file: figures/LAM_teaser.tex
% https://youtu.be/1zOoWSvPVxk?si=VG89O6cO14ZfEIm0
% Frames to be captured
% - man1 talking
% - people surprised 
% -        and happy 
% - man2 asking
% - people responding
% Audio 

\begin{figure}[tb]
\centering
    \begin{subfigure}[h]{0.69\linewidth}
        \includegraphics[width=\linewidth]{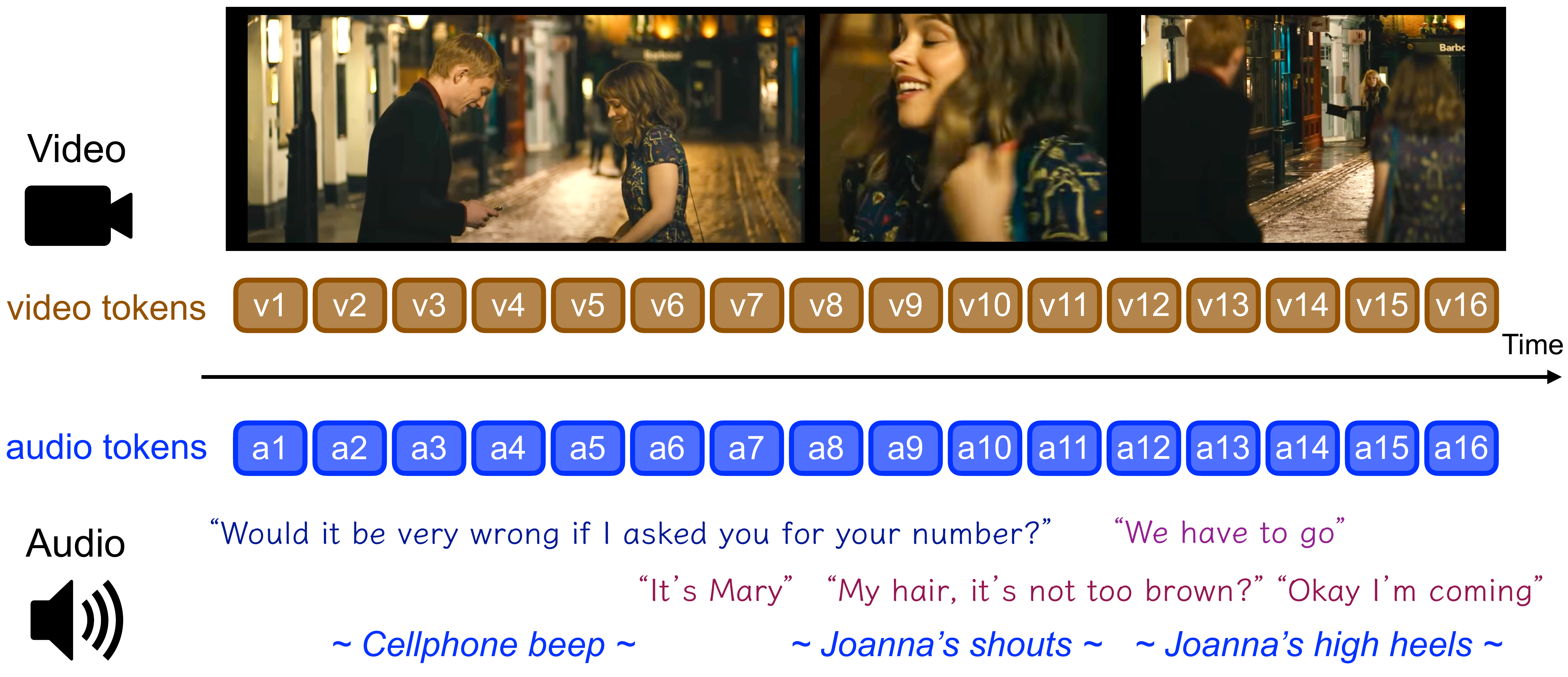}
        \caption{}
    \end{subfigure}
    \hfill
    \begin{subfigure}[h]{0.29\linewidth}
        \includegraphics[width=\linewidth]{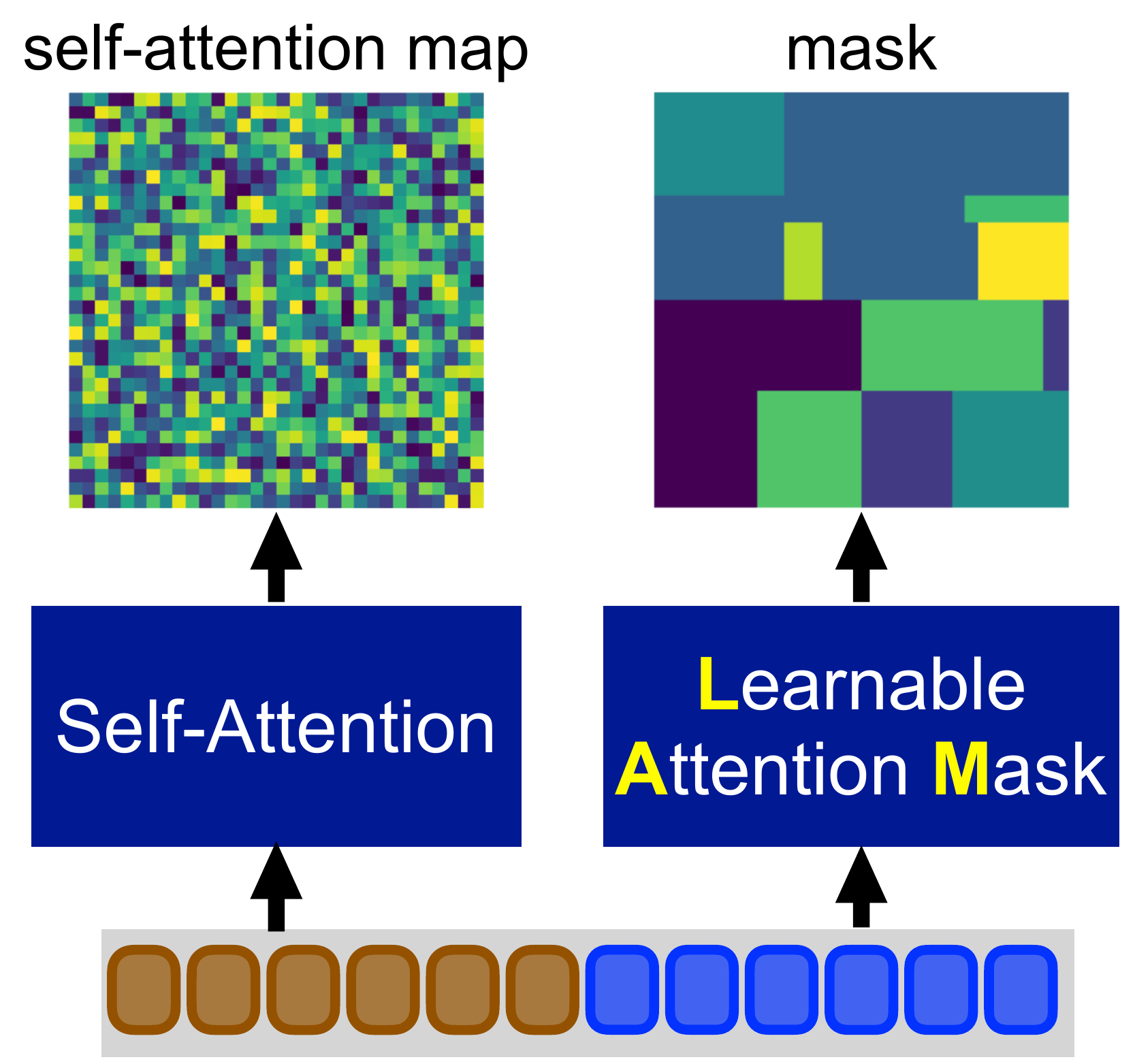}
        \caption{}
    \end{subfigure}
\caption{(a) While video and audio tokens naturally align in time, their associations can extend beyond temporal boundaries. For example, ``Joanna’s shouts'' may correspond to multiple video tokens (i.e. not just v8-11, but also v13-16). (b) The Self-Attention module~\cite{transformer} can capture these attention scores \textit{locally}, token-versus-token. We introduce the Learnable Attention Mask (LAM), a novel concept that enables a holistic overview of the entire sequence of input tokens, generating a mask that captures attention structures \textit{globally}.}
\label{fig:LAM_teaser}

\end{figure}

%% file: sections/2_related.tex
\section{Related Work}
\label{sec:related}

\subsection{Multimodal Transformers}
A predominant area of prior exploration in aligning multiple modalities centers around contrastive learning, a method extensively utilized in both image-text and video-audio alignment contexts~\cite{simclr,khosla2020supervised,clip,moco,han2023imagebindllm,zhang2023videollama}. Recent investigations have also delved into merging diverse modalities within a unified feature space through the incorporation of cross-attention layers~\cite{xattn,Lee2021CrossAttentionalAF,multixatt,qd_detr}. Furthermore, there is a growing trend of leveraging Transformer capabilities for multimodal fusion tasks~\cite{luo2021clip4clip, Kamath2021MDETRM, han2023imagebindllm,Barrios_2023_ICCV,moment-detr}. Our decision to employ a multimodal transformer in our design is rooted in its unparalleled capability to integrate information across diverse modalities, thus fostering a more comprehensive understanding of the input data. Through the utilization of this unified architecture, we are enabled to effectively capture intricate interactions within the sequence, strategically prioritizing relevant cues based on their significance. In contrast to conventional methodologies that treat modalities in isolation, the multimodal transformer facilitates the seamless integration of contextual information, thereby yielding more coherent and nuanced representations.

\subsection{Language Models for Video Description}
To adapt a Large Language Model (LLM) for AD generation, we incorporate an adapter module. This module processes audiovisual features and transforms them into the feature space of our LLM. The concept of training an adapter module rather than finetuning the entire LLM to account for a new modality has been widely explored~\cite{sung2022vladapter, hu2023llm}, but the method most similar to ours is LLaMA-Adapter~\cite{llama_adapterv1, llama_adapterv2}. LLaMA adapter, however, does not account for audio data. Our method follows that of LLaMA-Adapter closely, but changes the input feature space to include both audio and video features. The previous State-of-the-Art in our specific task (generating audio descriptions of movie clips) on the MAD dataset are the AutoAD models~\cite{autoad1,autoad2}. We are able to generate comparable results with significantly less fine tuning and contextual information. Recent models have also achieved significant results in finding important moments in longer videos, but these contributions are not particularly relevant to ours because we focus on describing shorter video segments~\cite{moment-detr, Barrios_2023_ICCV}. Another recent result similar to ours is the Video-LLaMA model, which focuses on general purpose visual question answering but uses a Q-Former instead of an adapter module to fuse the visual, audio, and text modalities~\cite{zhang2023videollama}.

%Additionally, Multi-modal Transformers are a family of deep learning models designed to handle complex, multi-modal data by combining the strengths of Transformer-based architectures~\cite{NIPS2017_7181} with multi-modal fusion techniques~\cite{luo2021clip4clip, radford2021learning, Kamath2021MDETRM}. This architecture family has achieved remarkable performance in various domains including computer vision ~\cite{Li2022SepViTSV,dosovitskiy2020image,li2022efficientformer,Arnab2021ViViTAV,Bertasius2021IsSA}, natural language processing~\cite{Radford2019LanguageMA,Brown2020LanguageMA,bert, brown2020language, wolf2020transformers}, and audio processing~\cite{gong21b_interspeech,verma2021audio}

%This approach makes sense for our use case because we do not need aligned feature spaces for audio and video, we just need a combined feature representation that effectively encodes both. Some alternatives which we did not explore is sharing a transformer between audio and video and averaging the outputs of each
%~\cite{Gong_2022, claborne2023behavior}, or using pretrained image-text models and fitting audio into the textual encoder
%~\cite{park2023clip}.

\subsection{Masking Attention}

In the field of Natural Language Processing, researchers have explored various methods of constructing attention masks, while also investigating their impact on transformer architectures~\cite{mask_networks,MDMAN,adapt_mask,self_mask}. Conversely, this exploration has received limited attention in Computer Vision~\cite{MST,swinbert}. Motivated by this disparity, our objective is to investigate this phenomenon, particularly in the context of multimodal data, and its implications for task performance. Unlike the approach proposed by SwinBert~\cite{swinbert}, which advocates for a sparse and learnable mask, our focus aligns more closely with the principles of Mask Attention Networks~\cite{mask_networks}. Instead of relying on a static mask matrix, which may restrict the model's ability to capture local relationships effectively, we propose employing a Learnable Attention Mask (LAM). This adaptive mechanism aims to prioritize and regulate attention tokens within long sequences based on their contextual significance in a dynamic manner.

%% file: sections/3_method.tex
\section{Learnable Attention Mask (LAM)}
\label{sec:lam}
Our goal is to train a Learnable Attention Mask that effectively prioritizes and regulates tokens based on their significance within a complex sequence. This adaptable mechanism can be seamlessly incorporated into any of the existing Transformer Encoders. Figure~\ref{fig:LAM_architecture} shows the overview of the LAM architecture. 

\input{figures/LAM_architecture}

\subsection{Definition}
We aim to generate an attention mask that adeptly prioritizes and regulates tokens according to their importance within a sequence. For this purpose, we develop a learnable module (denoted as $\mathcal{X}$), which receives a sequence of tokens $T$ as input and returns a mask $M$, as shown in Equation \ref{eq:mask}. However, the shape of the mask $M$ depends on the sequence length and the purpose of the multi-head attention, whether it's self-attention or cross-attention.
\begin{equation}
\label{eq:mask}
\mathcal{X}\left(T\right)\rightarrow{M}
\end{equation}
\textbf{Self-Attention.} In the context of self-attention, the resulting mask output can be represented as $L_t \times L_t$, where $L_t$ denotes the size of the input sequence $T$. Here, $\mathcal{X}$ corresponds to a Feedforward network (FFN).\\
\textbf{Cross-Attention.} In cross-attention,  $\mathcal{X}$  will still be associated with a Feedforward network. Nonetheless, the function's input is determined by the dot product of the query $Q$ and key $K$ tensors within the multi-head attention layer, as indicated in Equation~\ref{eq:mask_xatt}. Here, the shape of the mask $M$ is defined as $L_q \times L_k$, where $L_q$ is the length of the Query tensor and $L_k$ the length of the Key tensor.
\begin{equation}
\label{eq:mask_xatt}
\mathcal{X}\left(QK^T\right)\rightarrow{M}
\end{equation}
%\textbf{Attention Score.} The final attentions scores are computed following the Equation~\ref{eq:att_score}. This equation computes the attention scores by multiplying the masking matrix 
%$M$ to the scaled dot product of the query $Q$ and key $K$ matrices, then taking the softmax of the result and multiplying it by the value matrix $V$. The masking matrix is used to mask out certain positions in the input sequence to prevent the model from attending to them.

%\begin{equation}
%\label{eq:att_score}
%\text{Attention Scores} = \text{softmax} \left( \frac{QK^T * M_k}{\sqrt{d_k}} \right) V
%\end{equation}

\noindent\textbf{Scalability.} This mask can be applied either globally across all transformer layers in a stack or scaled individually for each transformer layer within the stack. This flexibility enables the use of different masks depending on the depth of the layers.
\subsection{Learnable Attention Mask Module}
Let $\mathbf{x} \in \mathbb{R}^{d_{\text{in}}}$ denote the input to the Learnable Attention Mask module, and $\mathbf{W}_i \in \mathbb{R}^{d{\text{in}_i} \times d{\text{out}_i}}$, $\mathbf{b}_i \in \mathbb{R}^{d{\text{out}_i}}$ be the weight matrix and bias vector of the $i$-th layer, respectively, where $i \in {1, 2, \ldots, L}$ and $L$ is the total number of layers in LAM. The forward pass of the Learnable Attention Mask module is denoted as follows:

\begin{equation}
\mathbf{h}_1 = \text{ReLU}(\mathbf{W}_1 \mathbf{x} + \mathbf{b}_1)
\label{eq:forward_pass1}
\end{equation}

\begin{equation}
\mathbf{h}_i = \text{ReLU}(\mathbf{W}_i \mathbf{h}_{i-1} + \mathbf{b}_i), \text{ for } i \in \{2, 3, ..., L-1\}
\label{eq:forward_pass2}
\end{equation}

\begin{equation}
M = \mathbf{W}_L \mathbf{h}_{L-1} + \mathbf{b}_L
\label{eq:forward_pass3}
\end{equation}

where $\text{ReLU}(\cdot)$ represents the rectified linear unit activation function, and $M$ denotes the generated mask. 

\subsection{Multi-Layer Learnable Attention Mask} \label{sec:multi_lam}
In the context of the $i^{th}$ layer of the self-attention Transformer, we denote the input sequence as $X^{(i)}$ and the resulting output as $Y^{(i)}$. Continuing with this notation, let's delve into the mathematical representation of the attention mechanism within each layer:
\begin{equation}
\text{Att}^{(i)} = \text{softmax}\left(\frac{Q^{(i)}(X^{(i)})K^{(i)}(X^{(i)})^\intercal}{\sqrt{d_k}} \odot M^{(i)}\right)
\end{equation}
% V^{(i)}(X^{(i)})
Where $Q^{(i)}$ and $K^{(i)}$ represent the query, and key matrices respectively for the $i^{th}$ layer. The mask $M^{(i)}$ is dynamically learned for each layer during training and adapts to the context within that layer, and $\odot$ is a element-wise multiplication. The output $Y^{(i)}$ of the $i^{th}$ layer serves as the input to the $(i+1)^{th}$ layer. Therefore, the attention mechanism changes for each layer due to the change in input. To show the change in mask and attention per layer, we need to consider the evolution of $X^{(i)}$ and $M^{(i)}$ as the layer index increases. The specific changes in attention and mask will depend on the architecture of the model and the learning process during training, with each layer potentially learning different attention patterns and mask distributions based on the evolving context of the input sequence.

\subsection{Learnable Attention Mask for Cross Attention} The Learnable Mask Attention module, crucial to the cross-attention mechanism, utilizes Query and Key vectors as fundamental input components. It calculates the dot product between the Query vector $Q$ and the transpose of the Key vector $K^T$ to assess their mutual interaction. This dynamic interplay guides the allocation of focus, enabling the model to intelligently prioritize information extraction and processing across both sequences. By evaluating the dot product of the Query and Key vectors, the model gains valuable insights into their nuanced relationship. This understanding empowers the model to intelligently prioritize attention, ensuring optimal information extraction and processing across both sequences.

%% file: figures/LAM_architecture.tex
\begin{figure}[ht]
\centering
\includegraphics[width=0.75\linewidth]{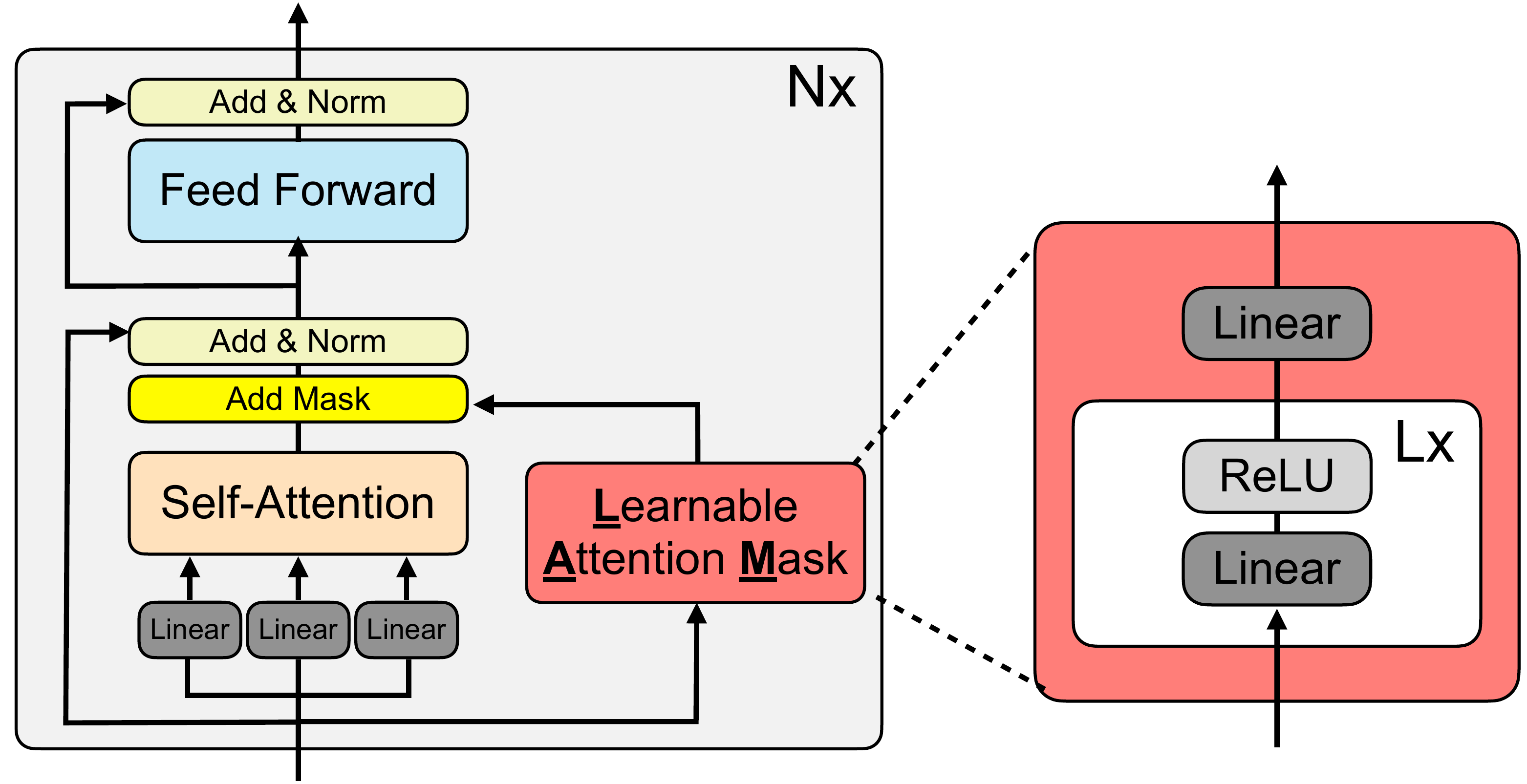}
\caption{Overview of the Learnable Attention Mask Architecture. The Learnable Attention Mask (LAM) module takes the entire sequence as input and generates a mask. This mask is then used for element-wise multiplication with the attention scores produced by the Transformer Encoders.}
\label{fig:LAM_architecture}
\end{figure}

%% file: sections/4_experiment.tex
\section{Experiments}

\subsection{Datasets}

\textbf{Generating AD.} MADv2~\cite{mad,autoad1} is a vast dataset for video-language grounding, with over $264$K queries in $488$ movies totaling $892$ hours. It includes MADv2-eval, with $10$ movies for evaluation.

\textbf{Moment Retrieval and Highlights Detection.} QVHighlights~\cite{moment-detr} is the latest dataset for moment retrieval and highlight detection, featuring annotations for both tasks in over 10,000 YouTube videos.

\textbf{Image Classification.} ImageNet 1K~\cite{imagenet} is a benchmark dataset consisting of $1.2$ million images across $1,000$ categories, commonly used for image classification tasks.

\textbf{Video Captioning Task.} MSRVTT~\cite{msrvtt} is a dataset for video captioning, comprising $10,000$ video clips from $20$ categories with human-annotated descriptions.

%\noindent\textbf{Generating AD.} MADv2~\cite{mad,autoad1} is a large-scale dataset for the video-language grounding task that comprises more than $264$K natural queries temporally grounded in $488$ full-length movies for a total of over $892$ hours of video. This dataset includes a subset named MADv2-eval, which comprises $10$ movies designated for evaluation purposes.

%\noindent\textbf{Moment Retrieval and Highlights Detection.} QVHighlights~\cite{moment-detr}, which is currently the most recently published dataset for moment retrieval and highlight detection, is also the sole dataset that includes annotations for both tasks. This dataset comprises more than 10,000 YouTube videos that have been annotated with text queries written by humans.

\subsection{Metrics}
%\noindent\textbf{Generating AD.} We employ conventional metrics to compare the generated Audio Description in text (AD) to the ground-truth AD. These metrics include Rouge-L~(R-L)~\cite{rouge}, CIDEr~(C)~\cite{cider} and Retrieval-based metric (R@k/N)~\cite{autoad2}. These metrics are less affected by the low-level variations in the test. A higher metric value signifies superior text generation when compared to the corresponding ground-truth.

%\noindent\textbf{Moment Retrieval and Highlights Detection.} The preferred evaluation metrics for video grounding tasks are Recall@$K$ and mAP@$K$ for IoU=$\theta$ (R@$K$-IoU=$\theta$), which assess both ranking and temporal overlap between predictions and annotations. In this study, models are evaluated using $K=1$ and IoU thresholds $\theta$ of 0.5 and 0.7. Additionally, average mAP across IoU thresholds from 0.5 to 0.95 with 0.05 increments is calculated, following a similar approach to action detection. Highlight detection is primarily assessed using mAP, while HIT@1 is employed to calculate the hit ratio for the highest scored clip, considering a clip positive if it receives a score of ``Very Good'', as per the methodology in previous works such as \cite{moment-detr,qd_detr}.

\textbf{Generating AD.} Conventional metrics like Rouge-L (R-L)\cite{rouge}, CIDEr (C)\cite{cider}, and Retrieval-based metric (R@k/N)~\cite{autoad2} are employed to compare generated Audio Descriptions (AD) with ground-truth AD. These metrics are robust to low-level variations in testing data, with higher values indicating superior text generation.

\textbf{Moment Retrieval and Highlights Detection.} For video grounding tasks, evaluation metrics include Recall@$K$ and mAP@$K$ for IoU=$\theta$ (R@$K$-IoU=$\theta$), assessing both ranking and temporal overlap. Models are evaluated at $K=1$ with IoU thresholds of 0.5 and 0.7. Average mAP across IoU thresholds from 0.5 to 0.95 with 0.05 increments is calculated. Highlight detection primarily employs mAP, while HIT@1 measures the hit ratio for the highest scored clip.

\textbf{Video Captioning.} Evaluation metrics for video captioning include BLEU4 (B4)~\cite{bleu}, CIDEr (C), SPICE (S)~\cite{spice}, METEOR (M)~\cite{meteor}, and Rouge-L (R-L), capturing different aspects of caption quality such as n-gram overlap, semantic similarity, and linguistic fluency.

\textbf{Image Classification.} Performance in image classification is often measured using Accuracy top-1 (Acc-top1) and Accuracy top-5  (Acc-top5).

\subsection{Baselines}
We integrated our contribution into four baseline models: LlaMA AdapterV2~\cite{llama_adapterv2} with a transformer-based audiovisual encoder, QD-DETR~\cite{qd_detr}, SwinBERT~\cite{swinbert}, and ViT Base~\cite{mae,transformer}. Our module, described in Section~\ref{sec:lam}, was added \textbf{only} to the encoder of each model, except for SwinBERT, which follows its design principle by replacing the fixed learnable mask with our approach. For more details, see supplementary material.
\input{tables/04_experiment_table_1_final}
%\noindent\textbf{Generating AD.} When it comes to generating AD, our study focuses on four model baselines: ClipClap~\cite{clipcap}, CapDec~\cite{capdec}, AutoAD-I~\cite{autoad1}, and AutoAD-II~\cite{autoad2}. These models serve as the foundation for our investigation into the efficacy and performance of AD generation.

%\noindent\textbf{Video Grounding.} We adopt various baselines, encompassing both proposal-based and proposal-free approaches, to comprehensively evaluate our method. The selected baselines for this particular task comprise CAL~\cite{cal}, XML~\cite{xml}, XML+\cite{xml}, Moment-DETR\cite{moment-detr}, and QD-DETR~\cite{qd_detr}. 
\subsection{Results}
To examine the impact of our method, we evaluate its performance across five different tasks as presented in Table~\ref{tab:final_results}. Our analysis reveals that our method yields substantial improvements when applied to multimodal encoders compared to single-modality encoders. For instance, in Table~\ref{tab:ad}, we observe a maximum improvement of $12.7$ for the R@$5/16$ metric, with an average improvement of $8.23$ across all metrics. Similar trends are evident in Tables~\ref{tab:mr_qv} and~\ref{tab:hd_qv}, where maximum improvements of $2.46$ and $0.93$, respectively, are observed alongside average improvements of $2.21$ and $0.86$, respectively. Conversely, when evaluating single-modality encoders, we generally observe minimal gains and occasional decreases in performance across certain metrics. For instance, in Table~\ref{tab:msrvtt}, we note a decrease of $0.79$ for the B4 metric, while slight increases of $0.28$ and $0.39$ are observed for the C and S metrics, respectively. Similarly, in Table~\ref{tab:imagenet}, a small improvement of $0.74$ is observed for Acc-Top1.

\textbf{Takeaway.} Our method leverages multimodal sequences more effectively than single-modality sequences

\input{tables/04_experiment_table_3}

\subsection{Ablation Study}

\noindent We investigate three crucial aspects of our Learnable Attention Mask (LAM) module's design: \textbf{(i)} The representation capacity and depth of the attention mask itself, and how the depth of representing the mask impacts model performance. \textbf{(ii)} Whether the performance gains are primarily attributed to LAM's masking and selective attention capabilities, or simply from having more trainable parameters. \textbf{(iii)} The optimal way to incorporate and fuse the learnable mask with the original attention map within the architecture.
Through our analysis, we uncover design principles for effectively integrating LAM into the overall model to maximally leverage its selective attention capabilities. Understanding the interplay between attention mask representation depth, masking fusion strategies, model performance, and architectural factors provides insights into maximizing LAM's efficacy for performance improvements.

%In this section, our focus lies in exploring the impact of our Learnable Attention Mask on the overall architecture. By doing so, we aim to uncover the design principles necessary for achieving optimal solutions. To accomplish this, we must address two fundamental questions: \textbf{(i)} How crucial is the accurate representation of the attention mask, and to what extent does it affect the model's performance?; and \textbf{(ii)} When observing performance improvements, is it primarily due to the addition of more trainable parameters, or is it attributed to the influence of the Learnable Attention Mask module itself?. We aim to understand how the attention mask, model performance, and architectural improvements are interconnected through thorough analysis and experiments. This exploration will provide valuable insights into the mechanisms that determine the effectiveness of our approach.

\noindent\textbf{Attention Mask Influence.} We analyze the performance impact of the Learnable Attention Mask module using a subset of MAD-v2. We measure performance with Rouge-L and CIDEr metrics across four experiments: (i) full attention, (ii) learnable sparse mask from SwinBert, (iii) our Learnable Attention Mask (LAM), and (iv) Multi-Layer LAM. Employing LAM significantly improves performance over the full attention baseline, increasing CIDEr from $15.46$ to $16.58$. Multi-Layer LAM achieves the best performance at $17.11$ CIDEr. In contrast, SwinBERT's sparse learnable mask drops performance from $15.46$ to $9.72$ CIDEr due to its inability to capture the dynamic nature of MADv2 scenes with shot changes, transitions, and soundtracks. Our LAM better handles such multimodal sequences.

\input{tables/04_experiment_table_4}

\noindent\textbf{Performance: Parameters vs. LAM Module Influence.} We investigate whether the performance gains from our Learnable Attention Mask (LAM) module stem from the module itself or simply from increased model parameters. We compare a baseline model with full attention against a multi-layer LAM model and a variant model with full attention but additional linear layers to match the LAM model's parameter count. Results on the audio description generation task (Table \ref{tab:params_ad}) show that merely increasing parameters without LAM leads to a drop in performance from $15.46$ to $12.87$ CIDEr. This suggests LAM's efficacy comes from its ability to selectively emphasize token attention, not just added parameters.

%In this section, we explore the impact of incorporating a Learnable Attention Mask (LAM) or increasing the parameters in a Transformer model. The inclusion of the Learnable Attention Mask module inherently expands the parameters within each Transformer layer. However, the key question revolves around whether the performance boost originates from the LAM module itself or simply from the augmented parameters. To investigate this, we conducted an experiment comparing a baseline model with full attention (no mask) against a multi-layer LAM configuration. Additionally, we introduced a variant model with full attention but integrated linear layers to adjust attention scores, thereby equating the parameter count to that of the LAM module. The performance analysis, as depicted in Table \ref{tab:params_ad}, focuses on the AD generation task within our validation subset. Interestingly, the augmentation of parameters does not translate to performance improvement; rather, there is a decline from $15.46$ to $12.87$ CIDEr (C). This discrepancy leads us to conclude that the efficacy of our LAM module is not solely dependent on the increased parameter count, but rather on its ability to capture nuanced notions and selectively emphasize token attention across the sequence.

% \input{figures/6_appendix_fig2}
% \input{figures/4_experiment_fig_ablation}
%\subsection{Qualitative Results}
\input{figures/6_appendix_fig3}

\input{figures/6_appendix_fig4}
\noindent\textbf{More Ablation Studies.} In the supplementary material, you will find an analysis of how the depth (L number) impacts the performance of the LAM module. Additionally, it explores the effect of using two element-wise operations for masking fusion and how LAM modifies the distribution of attention weights.

\section{Qualitative Analysis}
The qualitative analysis portrayed in Figure~\ref{fig:qualitative_analysis} provides insights into our implementation of Multi-Learnable Attention Mask (LAM) in the Audio Description generation task. The visual depiction highlights two aspects: Figure~\ref{fig:qual_sub1} shows the simultaneous audio and video signals, while Figure~\ref{fig:qual_sub2} illustrates the mask values corresponding to each token in the initial transformer layer. Despite slight zoom adjustments, the visual composition remains static, depicting the backyard of a house throughout all frames. Consequently, the LAM module activates only three visual tokens out of twenty-five, assigning minimal attention to the audio tokens. In contrast, the auditory elements exhibit captivating dynamics, transitioning from the sounds of wind, insects, and a frog to the rhythmic ticking of a clock. The ground truth Audio Description is: ``A set of swings and a climbing frame stand in a rural backyard, along with a picnic table and a brick barbecue".
The model prioritizes attention to outdoor sounds (wind, insects, frog) over indoor sounds (watch ticking), aligning with visual token attention. Initial outdoor audio tokens strongly attend to first visual backyard tokens. Less attention is given to tokens related to transitioning indoors, coherent with describing the rural backyard. To compare with self-attention, Figure~\ref{fig:distr_attention} shows attention weight distributions for Learnable Attention Mask (LAM) and full-attention on the same scene. Without LAM, the distribution is more uniform, suggesting more tokens receive attention. With LAM, the distribution is skewed left with many weights near zero, implying focused attention on fewer tokens. This analysis highlights LAM's capability to discern and prioritize specific tokens, enhancing multimodal scene interpretation.

%% file: tables/04_experiment_table_1_final.tex
\begin{table}[h]
\centering
\footnotesize
\begin{minipage}{\textwidth}
    \centering
    \begin{subtable}[h]{0.45\textwidth}
        \centering
        \begin{tabular}{@{}lcccc@{}}
            \toprule
            \textbf{Model} & \textbf{R-L} & \textbf{C} & \textbf{R@5/16}\\
            \midrule
            LlaMA~\cite{llama,llama_adapterv2}& $10.7$ & $9.4$ & $43.4$ \\ 
             \textbf{Ours} & $\mathbf{13.5}$ & $\mathbf{18.6}$ & $\mathbf{56.1}$\\ \midrule
            \textbf{Gain($\Delta$)} & $2.8$ & $9.2$ & $12.7$\\
            \bottomrule
        \end{tabular}
        \caption{AD Task on MADv2-named~\cite{mad,autoad1}}
        \label{tab:ad}
    \end{subtable}%
    \hfill
    \begin{subtable}[h]{0.45\textwidth}
        \centering           
        \begin{tabular}{@{}lcc@{}}
            \toprule
            \textbf{Model} & \textbf{R1@IoU0.7} & \textbf{mAP (Avg)} \\
            \midrule
            QD-DETR~\cite{qd_detr} &$44.98$ &$39.86$ \\ 
            \textbf{Ours} & $\mathbf{46.94}$ & $\mathbf{42.32}$ \\ \midrule
            \textbf{Gain($\Delta$)} & $1.96$ & $2.46$ \\
            \bottomrule
        \end{tabular}
        \caption{Moment Retrieval Task in QVHighlights~\cite{moment-detr}}
        \label{tab:mr_qv}
    \end{subtable}
    
    \vspace{0.1cm} % Adjust the space between the two groups of subtables
    
    \begin{subtable}[h]{0.45\textwidth}
        \centering
        \begin{tabular}{@{}lcc@{}}
            \toprule
            \textbf{Model} & \textbf{mAP} & \textbf{HIT@1} \\
            \midrule
            QD-DETR~\cite{qd_detr} &$38.94$ &$62.40$ \\ 
            \textbf{Ours} & $\mathbf{39.70}$ & $\mathbf{63.33}$ \\ \midrule
            \textbf{Gain($\Delta$)} & $0.76$ & $0.93$ \\
            \bottomrule
        \end{tabular}
        \caption{Highlights Detection at VeryGood confidence in QVHighlights~\cite{moment-detr}}
        \label{tab:hd_qv}
    \end{subtable}%
    \hfill
    \begin{subtable}[h]{0.45\textwidth}
        \centering
        \begin{tabular}{@{}lcc@{}}
            \toprule
            \textbf{Model} & \textbf{Acc-Top1} & \textbf{Acc-Top5} \\
            \midrule
            *ViT Base~\cite{mae,transformer} & $82.71$ & $96.32$ \\
            \textbf{Ours}  & $\mathbf{83.45}$ & $\mathbf{96.59}$ \\ \midrule
            \textbf{Gain($\Delta$)}  & $0.74$ & $0.27$ \\
            \bottomrule
        \end{tabular}
        \caption{Image Classification in ImageNet 1K~\cite{imagenet}}
        \label{tab:imagenet}
    \end{subtable}
    \vspace{0.1cm} % Adjust the space between the two groups of subtables

    \begin{subtable}[h]{0.45\textwidth}
        \centering
        \begin{tabular}{@{}lccccc@{}}
            \toprule
            \textbf{Model} & \textbf{B4} & \textbf{R-L} & \textbf{M} & \textbf{C} & \textbf{S} \\
            \midrule
            SwinBERT~\cite{swinbert} & $\mathbf{42.82}$ & $\mathbf{62.06}$ & $30.39$ & $51.96$ & $7.64$\\ 
            \textbf{Ours} & $42.03$ & $62.05$ & $\mathbf{30.60}$ & $\mathbf{52.24}$ & $\mathbf{8.03}$\\ \midrule
            \textbf{Gain($\Delta$)} & $-0.79$ & $-0.01$& $0.21$ & $0.28$ & $0.39$\\
            \bottomrule
        \end{tabular}
        \caption{Video Captioning Task in MSRVTT~\cite{msrvtt}}
        \label{tab:msrvtt}
    \end{subtable}

\end{minipage}
\caption{\textbf{Comparing performance across various datasets.} We evaluate our masking method on both multimodal encoders and single modality encoders. Our method demonstrates significant performance gains when applied to multimodal encoders, particularly in tasks (a, b, and c). However, for tasks involving single-modality encoders (d and e), we observe minimal improvements across most metrics. The asterisk (*) denotes that we retrained using the codebase and observed a slight decrease in performance compared to the numbers reported in~\cite{mae}.}
\label{tab:final_results}
\end{table}

%% file: tables/04_experiment_table_3.tex
\begin{table}[ht]
\centering
\begin{tabular}{@{}clclc@{}}
\toprule
\textbf{Mask}                   &  & \textbf{R-L} &  & \textbf{C} \\ \midrule
Full Attention                  &  & $12.92$        &  & $15.46$      \\
Learnable Attn Mask (Fixed)*    &  & $10.02$        &  & $9.72$       \\
Learnable Attn Mask             &  & $13.10$        &  & $16.58$      \\
Multi-Layer Learnable Attn Mask &  & $\mathbf{14.28}$        &  & $\mathbf{17.11}$      \\ \bottomrule
\end{tabular}
\vspace{0.2cm}
\caption{\small{\bf Attention Mask Influence.}  
We provide an analysis of the performance regarding the Audio Description generation task. Specifically, we utilized a subset comprising 1010 instances from the MADv2 (Supplementary Material). Our findings reveal an improvement in performance upon the integration of the Learnable Attention Mask module. Furthermore, we observed a slightly more pronounced enhancement with the incorporation of dynamic masks in each transformer layer (referred to as Multi-Layer Learnable Attention Mask).`*' means we follow the learnable sparse mask design in~\cite{swinbert}. Experiments were trained using the same hyper-parameters and 10 epochs. }
\label{tab:att_mask}
%\vspace{-0.8cm}
\end{table}

%% file: tables/04_experiment_table_4.tex
\begin{table}[ht]
\centering
\begin{tabular}{@{}ccc@{}}
\toprule
\textbf{Experiment}                   & \textbf{R-L} & \textbf{C} \\ \midrule
Baseline                              & 12.92        & 15.46      \\
Full Attention w/ same number params. & 11.23        & 12.87      \\
Multi-Layer Learnable Attn Mask       & \textbf{14.28}       & \textbf{17.11}      \\ \bottomrule
\end{tabular}
\vspace{0.2cm}
\caption{\small{\bf Performance: Parameters vs. LAM Module Influence.} We present the performance results of three models: the baseline model without masking, a Full Attention Transformer with an equivalent number of parameters to the Multi-Layer Learnable Attention Mask, and the Multi-Layer Learnable Attention Mask itself. Generally, our findings suggest that the number of parameters does not directly correlate with performance gains.
  }
%\vspace{-0.5cm}
\label{tab:params_ad}
\end{table}

%% file: figures/6_appendix_fig3.tex
\begin{figure}[ht!]
    \centering
    \begin{subfigure}[b]{1.0\textwidth} % Doubled the width
        \centering
        \includegraphics[width=\textwidth]{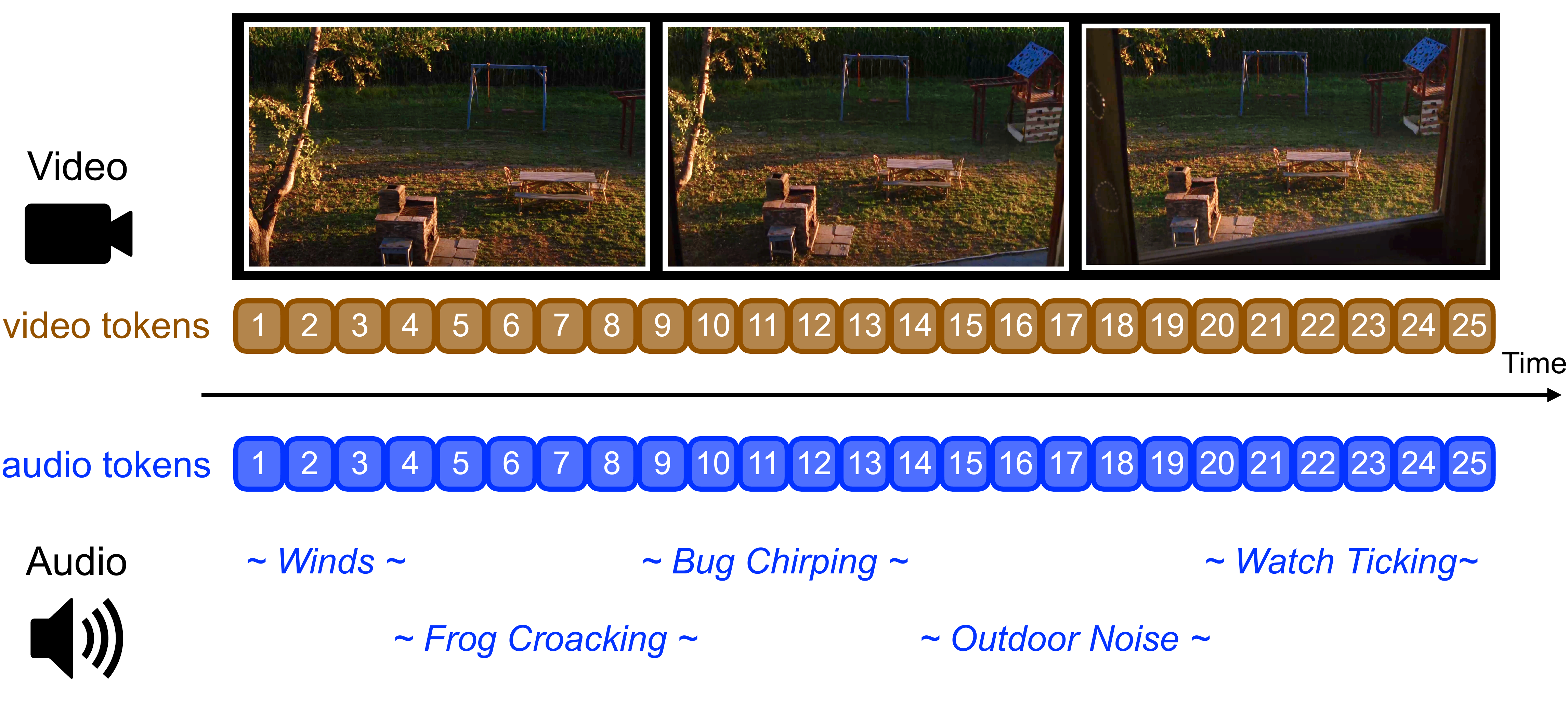}
        \caption{\textbf{Scene Visualization.} We highlight a specific moment from the movie Signs (2002) for qualitative analysis within the MADv2-eval set. Here, we meticulously present the visual elements while accurately representing the accompanying audio signals of the scene.}
        \label{fig:qual_sub1}
    \end{subfigure}
    \vskip\baselineskip
    \begin{subfigure}[b]{1.0\textwidth} % Doubled the width
        \centering
        \includegraphics[trim={0 0 1.5cm 0},clip,width=\textwidth]{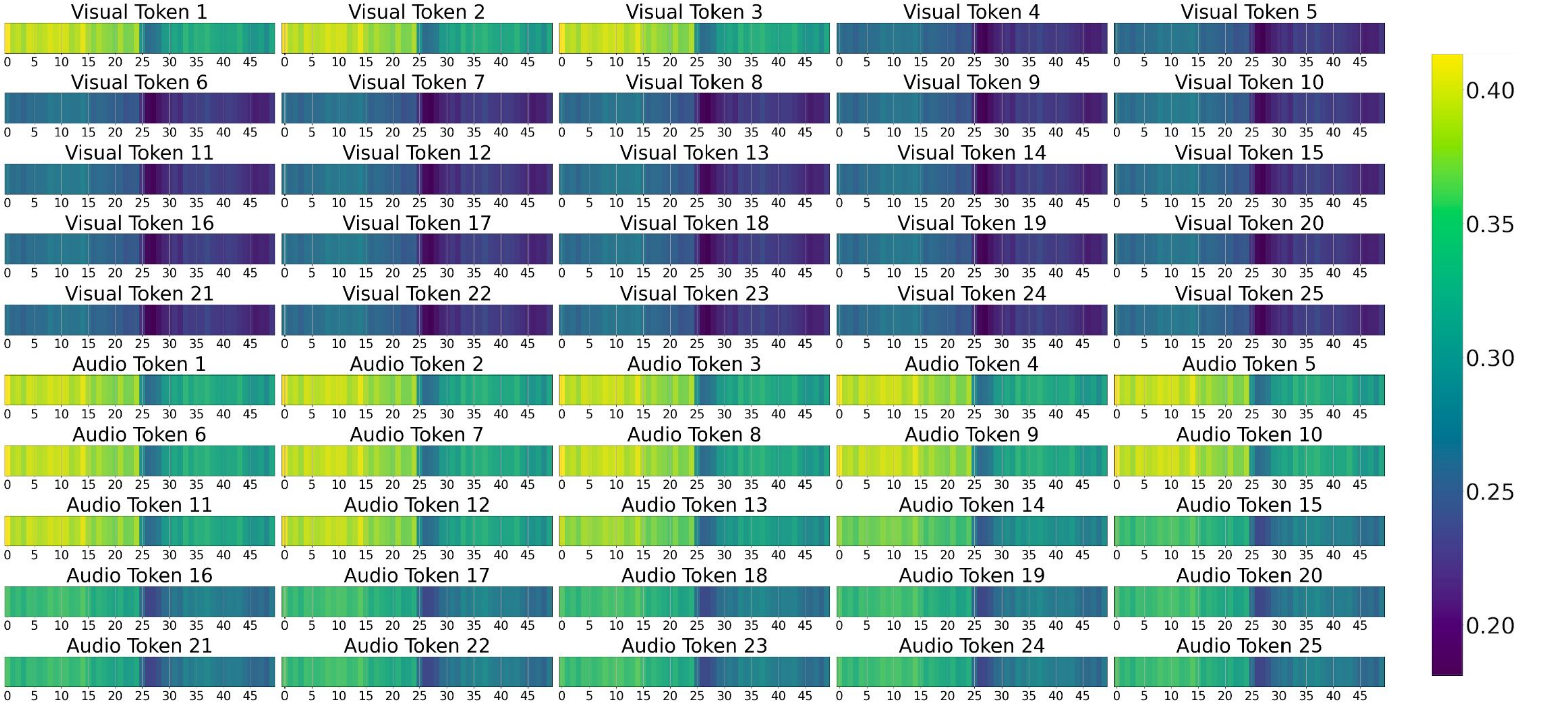}
       
        \caption{\textbf{Scene Visualization} We also showcase the mask values produced by the Learnable Attention Mask (LAM) module for each visual and audio token present in the scene. These mask values exhibit positive numerical values, ranging between 0 and 1 inclusively.}
        \label{fig:qual_sub2}
    \end{subfigure}
   \caption{\textbf{Qualitative Analysis.} This illustration presents a qualitative analysis of a specific instance from the MADv2-eval dataset. It depicts visual and audio signals alongside mask values corresponding to the initial transformer layer (1st layer). Video tokens are represented on the x-axis from 0 to 24, while audio tokens range from 25 to 50 on the same axis.}

    \label{fig:qualitative_analysis}

\end{figure}

%% file: figures/6_appendix_fig4.tex
\begin{figure*}[ht!]
    \centering
     %\vspace{-0.4cm}
    \includegraphics[width=0.55\textwidth]{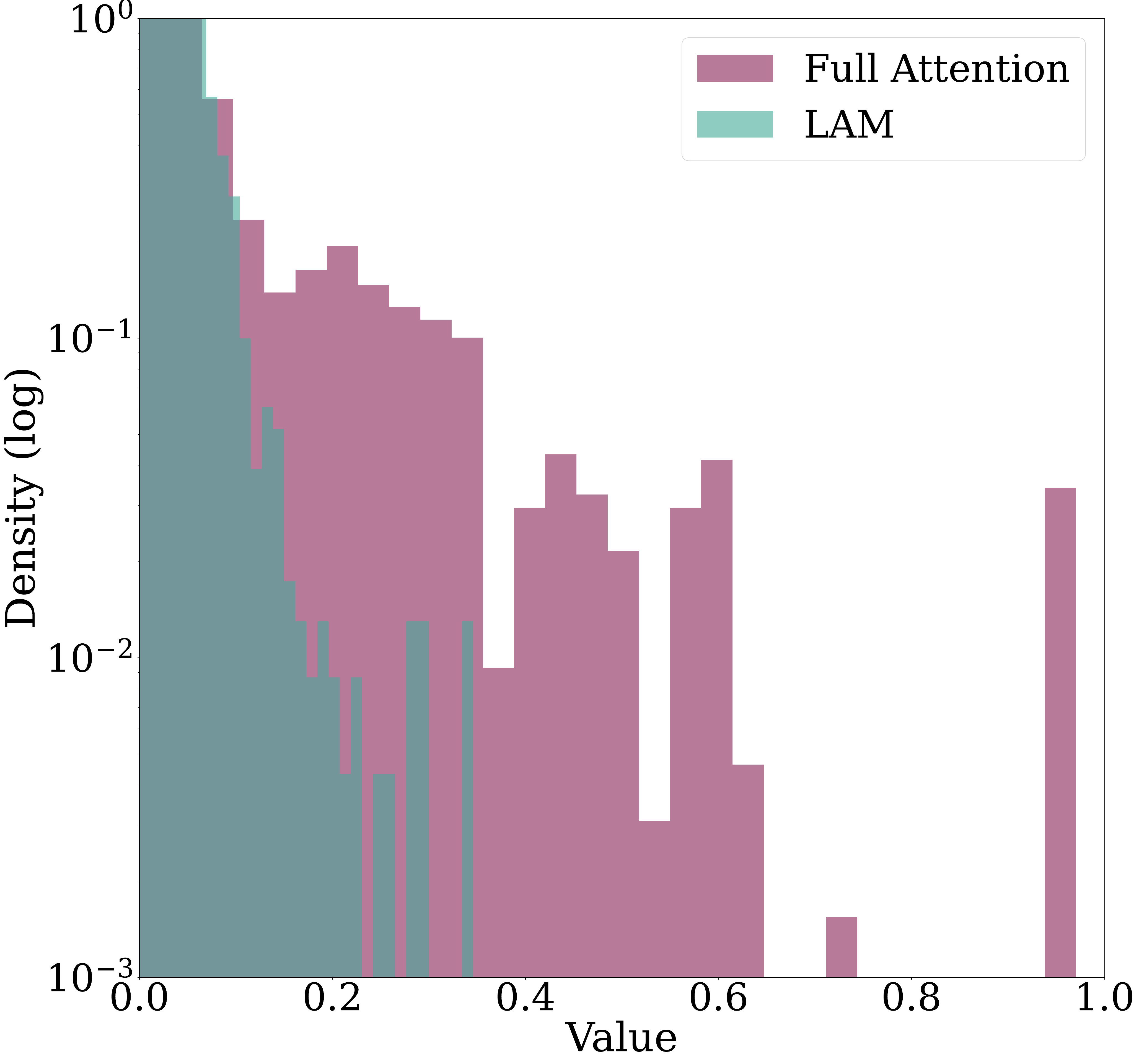}
    % \vspace{-0.3cm}
    \caption{\small\textbf{Analysis of Attention Weight Distribution in the Qualitative Example.} The plot illustrates the distribution of attention weights within the initial transformer layer across two distinct configurations: employing Learnable Attention Mask (LAM) and full-attention mechanisms. It is evident from the depiction that attention weights under LAM tend to exhibit a leftward bias, resulting in a significant portion approaching 0 or nearing zero. The distribution weights correspond to the same example in Figure~\ref{fig:qualitative_analysis}.
    }
    \label{fig:distr_attention}
    %\vspace{-0.5cm}
\end{figure*}

%% file: sections/5_conclusion.tex
\section{Conclusions and Limitations}
In this work, we introduce the Learnable Attention Mask (LAM) as a novel solution to tackle challenges in attention mechanisms. The LAM enables models to embrace a comprehensive perspective of the entire input, eventually regulating attention maps, prioritizing critical tokens, and reducing unnecessary computations. While our model demonstrates significant improvements in multimodal settings, we recognize the potential for further enhancement through the inclusion of additional contextual information during inference, such as character names for AD task. Additionally, we observe that the gains are not as significant in single-modality settings. Nevertheless, we believe that our proposed approach and experimental results will serve as a foundational stepping stone for future research in understanding and applying attention mechanisms across various scenarios.

% \textbf{Limitations.}\textcolor{red}{ 
% - Our model cannot predict characters names in MAD.
% - 
% }
% \noindent\textbf{Conclusion.}

%% file: sections/6_appendix.tex
\renewcommand\thesection{\Alph{section}}
\setcounter{section}{0}
\renewcommand{\thetable}{S\arabic{table}}  
\renewcommand{\thefigure}{S\arabic{figure}}

\section{Multimodal Encoder Tasks}
\label{sec:multi_task}
We evaluated the effectiveness of the learnable mask attention across three significant multimodal tasks: Audio Description (AD) Generation, Moment Retrieval, and Highlights Detection. In this section, we first provide details in these tasks (Sections~\ref{sec:task_ad} and~\ref{sec:task_highlight}) and show the application of our proposed LAM module to each task (Sections~\ref{sec:implementation_ad} and~\ref{sec:implementation_highlight}).

\subsection{Audio Description Generation}
\label{sec:task_ad}

Our task involves adapting a Large Language Model (LLM) to generate Audio Descriptions (AD) in text for a long-form movie $\mathcal{L}$ segmented into short clips $\{c_1, c_2, \ldots, c_N\}$. Each clip encompasses $\mathcal{S}$ samples in the visual stream (represented as $V$) and $S$ samples in the audio stream (denoted as $A$)\footnote{Raw sound from movies, excluding descriptions}. Specifically, our goal is to create a text $t_i$ that describes the audiovisual content presented in each clip $c_i$, aiming to assist individuals who are blind in following the movie's narrative.

\noindent \textbf{Audiovisual Model $\mathcal{AV}$.} We aim to train an audiovisual model that comprehends the relationships between sequential video and audio streams. Consequently, $\mathcal{AV}$ processes video ($V$) and audio ($A$) observations sampled at $c_i$ clip and produces an audiovisual feature representations $E_{va}$.

\begin{equation}
    \mathcal{AV}\left(V,A\right)\rightarrow E_{va}
\end{equation}
\\
\noindent \textbf{Large Language Model $\mathcal{H}$.}  Given an input sequence  $X = \{x_1, x_2, \ldots, x_n\}$, the model $\mathcal{H}$ estimates the probability distribution of the next word $x_{n+1}$ based on the context using the chain rule of probability:

\begin{equation}
P(x_{n+1} | X) = P(x_{n+1} | x_1, x_2, ..., x_n)
\end{equation}
The model is trained by maximizing the likelihood of generating the correct sequence according to the training data. During inference, it predicts the most likely next word given the context. The model's weights $\theta$ are optimized through back-propagation and gradient descent to improve its language understanding and generation capabilities.

\noindent \textbf{Adapter Module $\mathcal{P}$.} Let's assume a pre-trained model with parameters represented by $\theta$. The adapter layer introduces additional parameters for audiovisual understanding task, and these parameters can be denoted as $\phi$. The output of the adapter layers can be represented as $P(x',\phi)$, where $x'$ is the projected audiovisual features into the language space. So, the overall output of the model with the adapter layer can be written as:

\begin{equation}
     \mathcal{F}(\mathcal{H}(x, \theta), \mathcal{P}(x', \phi))\rightarrow t_i
\end{equation}

Where $\mathcal{F}$ is a function that combines the pre-trained Language Model $\mathcal{H}$ and the adapter $\mathcal{P}$ to produce an AD in text $t_i$.

\subsection{Moment Retrieval and Highlights detection}
\label{sec:task_highlight}

The visual-language grounding model, denoted as $\mathcal{G}$, is tasked with processing an untrimmed video, $V$, sampled from a temporal window $W$, along with a natural language query $Q$. It then produces predictions for $J$ temporal moments, defined as:

\begin{equation}
\label{eq:ground}
\mathcal{G}\left(V,Q\right)\rightarrow{(\tau_{s},\tau_{e}, s,s_{l})}^J_1.
\end{equation}

In Equation~\ref{eq:ground}, the grounding models yield a series of moments ranked by their confidence scores. Here, $(\tau_{s},\tau_{e})$ represents the duration span of the moment, while $s$ indicates its confidence score. Now, let's define the inputs for our attention modules. Given a video comprising $L$ clips and a text query containing $N$ words, their representations extracted by frozen video and text encoders are denoted as ${v_1, v_2, \ldots, v_L}$ and ${t_1, t_2, \ldots, t_N}$, respectively. Additionally, the grounding model provides saliency scores $s_{l}$ for each moment for the highlight detection task.

\subsection{Implementation Details for AD Generation}
\label{sec:implementation_ad}

\noindent\textbf{Feature Extraction.} The extraction of visual features follows the CLIP-based methodology outlined in~\cite{mad}. To be more specific, visual features are extracted at a rate of $5$ frames per second (FPS) with an embedding dimensionality of $D_v{=}512$. For audio feature extraction, we follow~\cite{Barrios_2023_ICCV} by utilizing the OpenL3~\cite{openl31,openl32} checkpoint pre-trained on videos containing environmental audiovisual data. We use a spectrogram time-frequency representation with $128$ bands and set the audio embedding dimensionality $D_a$ to $512$. Furthermore, we extract the audio embeddings using a stride size of $0.2$ seconds, \textit{i.e.}, with an extraction frame rate of $5$ Hz, matching the frame rate of the visual features.  

\noindent \textbf{Audiovisual Model $\mathcal{AV}$.} We utilize a Multimodal Transformer with a standard configuration~\cite{transformer}. For each observation $c_i$, consisting of both visual and audio information, we employ $S=25$ visual tokens and $S=25$ audio tokens, effectively spanning a $5$-second duration at a frame rate of $5$ FPS. This Multimodal Transformer architecture comprises $16$ layers and employs a Multi-Layer Learnable Attention Mask with a dimensionality of 768.
\looseness-1

\noindent \textbf{Large Language Model $\mathcal{H}$.} For Large Language Model, we choose to employ a frozen LLaMA $7$B model~\cite{llama} and opt to use its official checkpoint.

\noindent \textbf{Adapter Module $\mathcal{P}$.} We build our audiovisual adapter following the approach done in~\cite{llama_adapterv2}. In this part, we select $16$ tokens as audiovisual tokens. We adjust the last $31$ layers of LLaMA $7$B, making sure that the audiovisual features stay at a size of $512$, which then maps to $4096$ (LLaMA dimensionality). We set the depth to $8$, use $16$ heads, apply LoRA Rank~\cite{lora} with a value of $16$, and activate Bias layers~\cite{llama_adapterv1}. 

\noindent\textbf{Training Protocol.} To generate Audio Descriptions, we follow the training methodology outlined in ~\cite{llama_adapterv1,llama_adapterv2}. This involved utilizing $8$ RTX $6000$ Ada Generation GPUs, each equipped with $50$ GB VRAM, alongside employing a base learning rate of $1e-4$ and the Adam optimizer.

\subsection{Implementation Details for Moment Retrieval and Highlighting Task}
\label{sec:implementation_highlight}

\noindent\textbf{Feature Extraction.} The visual and text embeddings are extracted following the methodology presented in~\cite{moment-detr}. For video, we use SlowFast~\cite{SlowFastNF} and the visual encoder (ViT-B/32) of CLIP~\cite{clip} to extract features every 2 seconds. We then normalize the two features and concatenate
them at hidden dimension. The resulting visual features is denoted as $E_V \in \mathbb{R}^{L_{V} \times D_{V}}$, with $D_{V}=2816$. For text features , we use the CLIP text encoder to extract token level features, $E_V \in \mathbb{R}^{L_{Q} \times D_{Q}}$ with  $D_{V}=512$. %Next, we use separate 2-layer perceptrons with layer-norm and dropout to project the video and query features into a shared embedding space of size $d$.

\noindent\textbf{Video Grounding Model.}We adopt the methodology outlined in~\cite{qd_detr}. The architecture consists of three distinct components: an encoder comprising four layers of transformer blocks (two cross-attention layers and two self-attention layers), while the decoder has only two layers. We configure the hidden dimension of the transformers to be $256$  Additionally, for the transformer encoder layers and the cross-attention layers, we utilize our Learnable Mask Attention method as detailed in Section~\ref{sec:lam}.

\noindent\textbf{Training Protocol.} We conducted training over $200$ epochs, employing a batch size of 3$2$ and a learning rate set to $1e-4$. We utilized the Adam optimizer with a weight decay of $1e-4$, leveraging a single GPU, the RTX $6000$ Ada Generation.

\section{Single Modality Encoder Tasks}

\subsection{Image Classification Task}
\label{sec:task_classification}

In the image classification task, the goal is to assign an input image $I$ to one or more predefined classes from a set of $C$ classes. Let's denote the image classification model as $\mathcal{M}$. Given an input image $I$, the model generates a set of class predictions and their corresponding confidence scores:
\begin{equation}
\mathcal{M}(I) \rightarrow {(\hat{y}_1, \hat{p}_1), (\hat{y}_2, \hat{p}_2), \ldots, (\hat{y}_C, \hat{p}_C)}
\end{equation}
Here, $\hat{y}_c \in {1, 2, \ldots, C}$ represents the predicted class label for the $c$-th class, and $\hat{p}_c \in [0, 1]$ is the corresponding confidence score or probability assigned by the model to that class. The model's goal is to accurately predict the true classes present in the input image $I$.

\subsection{Video Captioning Task}
\label{sec:task_captioning}
In the video captioning task, the goal is to generate a textual description or caption for a given input video $V$. Let's denote the video captioning model as $\mathcal{M}$. Given an input video $V$, the model generates a sequence of words $W = {w_1, w_2, \ldots, w_N}$ that forms the caption:
\begin{equation}
\mathcal{M}(V) \rightarrow W = {w_1, w_2, \ldots, w_N}
\end{equation}
Here, each $w_i$ represents a word in the generated caption, and $N$ is the length of the caption sequence. The model's objective is to produce a natural language caption $W$ that accurately and coherently describes the content and events depicted in the input video $V$.

\subsection{Implementations Details for Image Classification Task}
We follow the pre-trained model developed in~\cite{mae} and fine-tune it for the image classification task. The base model is a Vision Transformer (ViT) with a $16$x$16$ patch size, $768$-dimensional embedding, $12$ transformer layers, and $12$ attention heads. It includes an MLP ratio of $4$, biases in the query, key, and value projections, and layer normalization with an epsilon of $1e-6$. To incorporate our proposed Learnable Attention Mask (LAM) module, we use the Multi-LAM variant, which generates the attention mask using a single linear layer without ReLU activation. For the pretraining stage, we adhere to the methodology outlined in \cite{mae}, but increase the batch size to $128$ and use $4$ gradient accumulation steps. For fine-tuning on the image classification task, we maintain a batch size of $128$ and $4$ gradient accumulation steps. Additionally, we train for $100$ epochs, apply a weight decay of $0.05$, set the drop path rate to $0.1$, and use mixup and cutmix with values of $0.8$ and $1.0$, respectively.

\subsection{Implementations Details for Video Captioning Task}

We adopt the methodology proposed by SwinBERT~\cite{swinbert}, with a notable modification. Instead of using a fixed learnable mask implemented via \texttt{nn.Parameter}, we integrate our Learnable Attention Mask (LAM) module, which consists of 16 layers while maintaining the same dimensionality as the original SwinBERT. Regarding the hyperparameters, the experiment utilizes a batch size of $2$ per GPU, running for $20$ epochs with a learning rate of $0.0003$. Training is conducted in half precision using DeepSpeed, with gradient accumulation over $16$ steps. The setup employs $8$ A$6000$ Ada generation GPUs, ensuring efficient and powerful computational performance.

\section{Ablation Study}

\input{figures/6_appendix_fig2}
\subsection{Influence of the Depth and Masking Fusion} 
In our experimental exploration of the cross-attention setup, we employ an ablation analysis to systematically assess the effects of removing two key components from the Learnable Attention Mask (LAM) module. These components encompass both the variation in the number of layers within the LAM module, as outlined in Section~\ref{sec:lam} of the main paper, and the distinct functionalities of mask operations: addition and multiplication. Figure~\ref{fig:able_qvh} presents a detailed analysis of the performance outcomes achieved through the Average mAP metric for the Moment Retrieval task on the QVHighlights validation dataset. Delving deeper into the results, it is apparent that the utilization of LAM (Learnable Attention Mask) contributes to notable enhancements in performance. These improvements vary across different configurations, with some showcasing more substantial gains than others. Nevertheless, regardless of the specific layer types employed or the operations conducted, the results consistently surpass the baseline performance. Remarkably, in this particular examination, the most favorable outcomes were attained by employing $32$ layers combined with both addition and multiplication operations. These configurations yielded impressive Average mAP metrics of $42.61$ and $42.32$ respectively, underscoring the efficacy of this approach in enhancing the performance of the Moment Retrieval task on the QVHighlights validation set.

\textbf{Takeaway.} While we mentioned in Section~\ref{sec:lam} of the main paper that the mask is integrated through element-wise multiplication, we also provide the option to incorporate addition at the element-wise level.

\input{figures/4_experiment_fig_ablation}
\subsection{Attention Weights.} Figure~\ref{fig:attn} illustrates the attention weight distribution across three layers in the Transformer Network: the $1$st, $8$th, and the last layer, under both full-attention settings and utilizing Multi-Layer Learnable Attention Mask (LAM). Notably, with our Multi-Layer LAM, a significant portion of attention weights diminishes to zero, while others converge towards values close to zero. This observation suggests that LAM has the potential to optimize the model training by minimizing redundant computations.

\section{Additional Details for Audio Description Generation}
\label{sec:exp_details}
In the following sections, we examine specific details that have not been addressed in the main paper. This comprehensive discussion includes insights into the current methodology for calculating metrics, the specific prompts employed, and the intricacies of both the training and evaluation processes for our implementation.

\subsection{Metrics} 

We employ the \texttt{pycocoeval} package from the \href{https://github.com/tylin/coco-caption}{coco-caption} repository to compute CIDEr~\cite{cider}. For CIDEr, we adopt the default parameters of $n=4$ and $sigma=6$, as outlined in~\cite{cider}. However, for Rouge-L~\cite{rouge}, a widely utilized metric in the Natural Language domain, we utilize a more convenient package available from Hugging Face. The Rouge-L metric can be accessed through the following link: \href{https://huggingface.co/spaces/evaluate-metric/rouge}{evaluate-metric/rouge}. When configuring Rouge-L, we set the input arguments as \texttt{use\_aggregator=True} and \texttt{use\_stemmer=True}, following the default setup. It's important to note that before computing the metrics, both the predicted text and the ground-truth text undergo transformation to lowercase and are stripped of punctuation.

To compute the Retrieval-based metric (R@k/N), we followed the methodology introduced in~\cite{autoad2}. Furthermore, this metric is complemented by BerScore~\cite{bertscore}, and for score consistency, we utilize the following hash code: \href{https://huggingface.co/spaces/evaluate-metric/bertscore}{\url{roberta-large_L17_no idf_version=0.3.12(hug_trans=4.30.2)-rescaled}}. This hash code specifies the BERT model version and the Hugging Face version.

\subsection{Natural Language Prompting}  

In order to enable the Audio Description functionality within our model, we employ the same prompting methodology as developed in LLaMA Adapter~\cite{llama_adapterv2}. As a result, the comprehensive prompt utilized for Audio Description generation is as follows: \textbf{``Below is an instruction that describes a task. Write a response that appropriately completes the request''}. Subsequently, we include the following instruction: \textbf{``Generate a caption for this video''.} The entire prompt is depicted in Figure \ref{code:prompt}.
\input{figures/6_appendix_fig1}

\subsection{Dataset Split}
As MADv2 lacks a validation set, we curated a subset of 1010 moments from two movies, \texttt{3034\_IDES\_OF\_MARCH} and \texttt{3074\_THE\_ROOMMATE} from the Unnamed version for our ablation studies and model selection. All models and experiments were assessed under consistent parameters to ensure fair comparisons. However, Table 1 in the main paper was generated using the entire dataset in the named version to maintain parity with other baselines.
\subsection{Training Protocol} 

The training procedure for our Audio Description Generation model strictly followed the prescribed protocol outlined in \cite{llama_adapterv2}. Initially, a preliminary training phase was conducted, with a sole focus on aligning the audiovisual features. This crucial step ensured the synchronization and coherence between the audio and visual elements of the input data. Subsequently, the model derived from this phase was utilized to resume training, with a specific emphasis on training the bias and gate layers as proposed in \cite{llama_adapterv2} for LLaMA \cite{llama} 7B architecture, in conjunction with the audiovisual encoder. In the following training phase, we did back-propagation the specified layers mentioned above. This was aimed at improving the model's performance in generating Audio Descriptions.

The entire training process spanned over 20 epochs, during which the best model in our validation subset was selected. This process involved the meticulous selection of hyperparameters, including a learning rate of $1e-4$, a weight decay of $0.05$, and a batch size of $256$. The AdamW optimizer was employed to effectively optimize the model parameters. During the alignment process, we train the audiovisual and adapter layers for $2$ epochs while keeping the remainder frozen to facilitate smoother convergence and alignment of the audiovisual features. It is noteworthy that throughout the entire training regimen, the LLaMA model remained frozen, thereby preserving its integrity and preventing inadvertent alterations to its pre-trained parameters.\\

\input{figures/6_appendeix_qualitative}
\section{Qualitative Analysis}

Figure~\ref{fig:failure_analysis} illustrates a failure case of our Learnable Attention Mask (LAM) method. In this example, we use the same visual information as Figure~\ref{fig:qualitative_analysis} of the main paper, but the audio samples correspond to $25$ tokens from the credits section, containing only \textit{soundtrack music}. Although LAM successfully assigns minimal values to the visual tokens, indicating its ability to recognize the lack of relevance between the video and audio tokens, it fails to handle the audio tokens appropriately. Instead of assigning values close to zero for the audio tokens, which would be the desired behavior, LAM assigns intermediate values from the distribution, suggesting a misinterpretation of the audio information's significance.

%This analysis highlights the remarkable ability of the LAM model to discern and prioritize specific tokens, thereby enhancing its proficiency in interpreting complex scenes through the utilization of multimodal cues.

%% file: figures/6_appendix_fig2.tex
\begin{figure*}[ht]
    \centering
     %\vspace{-0.4cm}
    \includegraphics[width=0.75\textwidth]{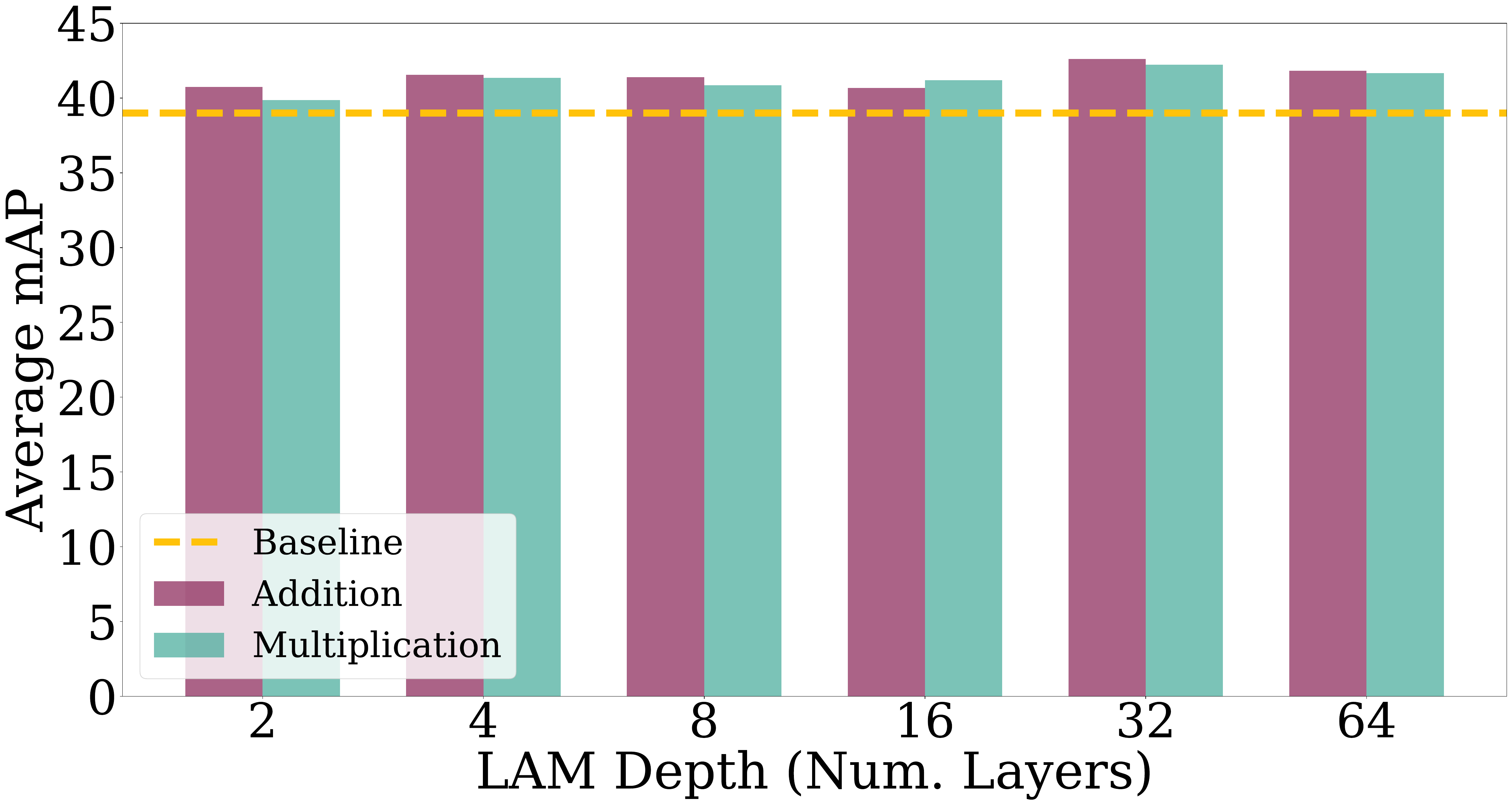}
    % \vspace{-0.3cm}
    \caption{\small\textbf{Ablation Studies on the number of layers in LAM and types of mask operation.} We conduct an investigation into the impact of varying the number of layers utilized within the Learnable Attention Mask (LAM) framework, as applied in the cross-attention configuration, along with the methods employed for mask fusion with attention weights. The experimentation involves the manipulation of the number of layers, ranging from $2$ to $64$, and explores two distinct fusion techniques: multiplication and addition operations, both implemented at the element-wise level. Evaluation of these experiments is carried out on the validation split set of QVHighlights~\cite{moment-detr}. Overall, notable enhancements in performance, particularly concerning the Average mAP metric for the Moment Retrieval task, are observed. The most substantial improvements are achieved when utilizing $32$ layers within the LAM module.
    }
    \label{fig:able_qvh}
    \vspace{-0.2cm}
\end{figure*}

%% file: figures/4_experiment_fig_ablation.tex
\begin{figure}[ht]
%\centering
\begin{subfigure}{0.45\linewidth}
\includegraphics[width=\linewidth]{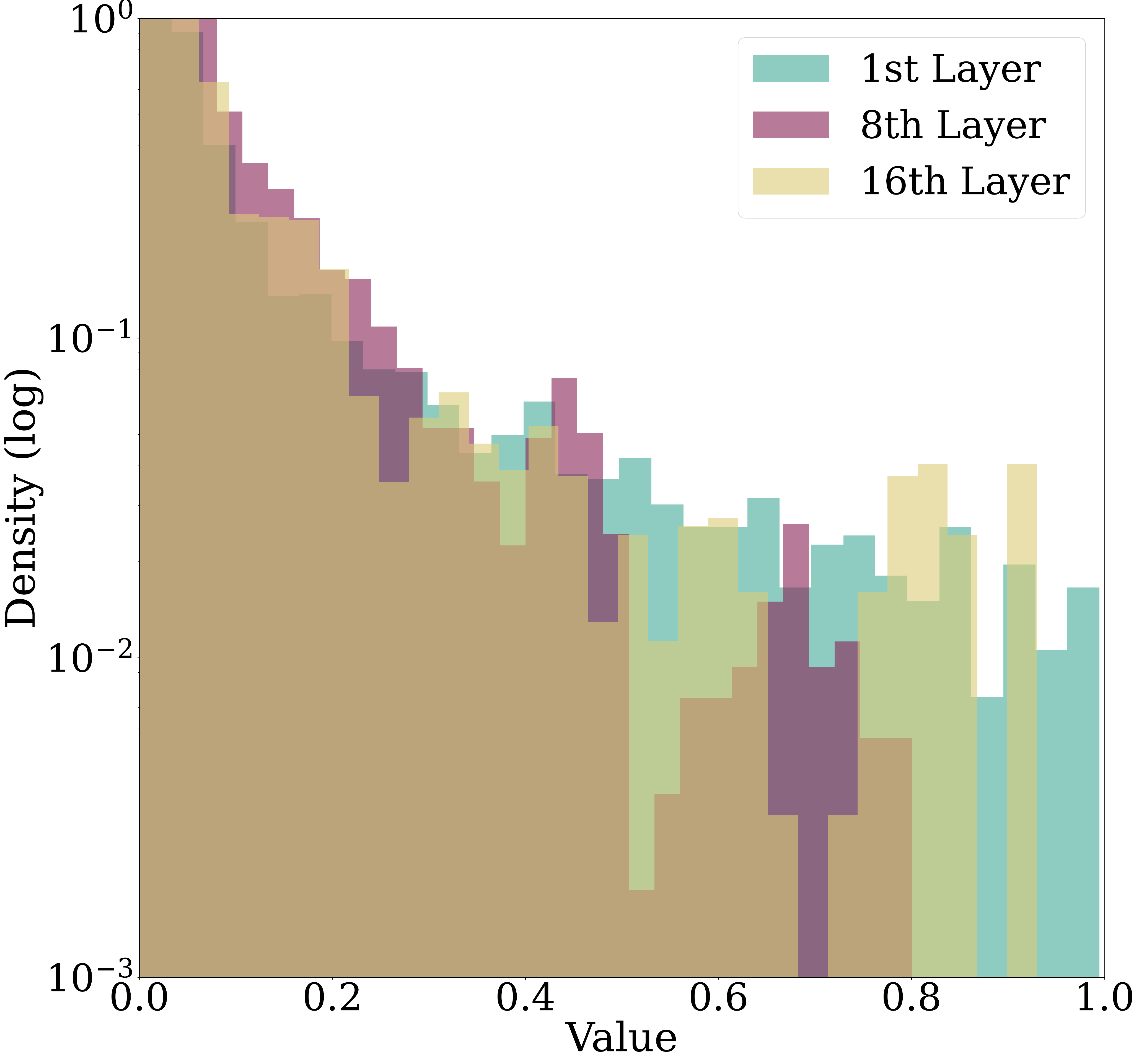}
\caption{Using Full Attention}
\end{subfigure}
\hfill
\begin{subfigure}{0.45\linewidth}
\includegraphics[width=\linewidth]{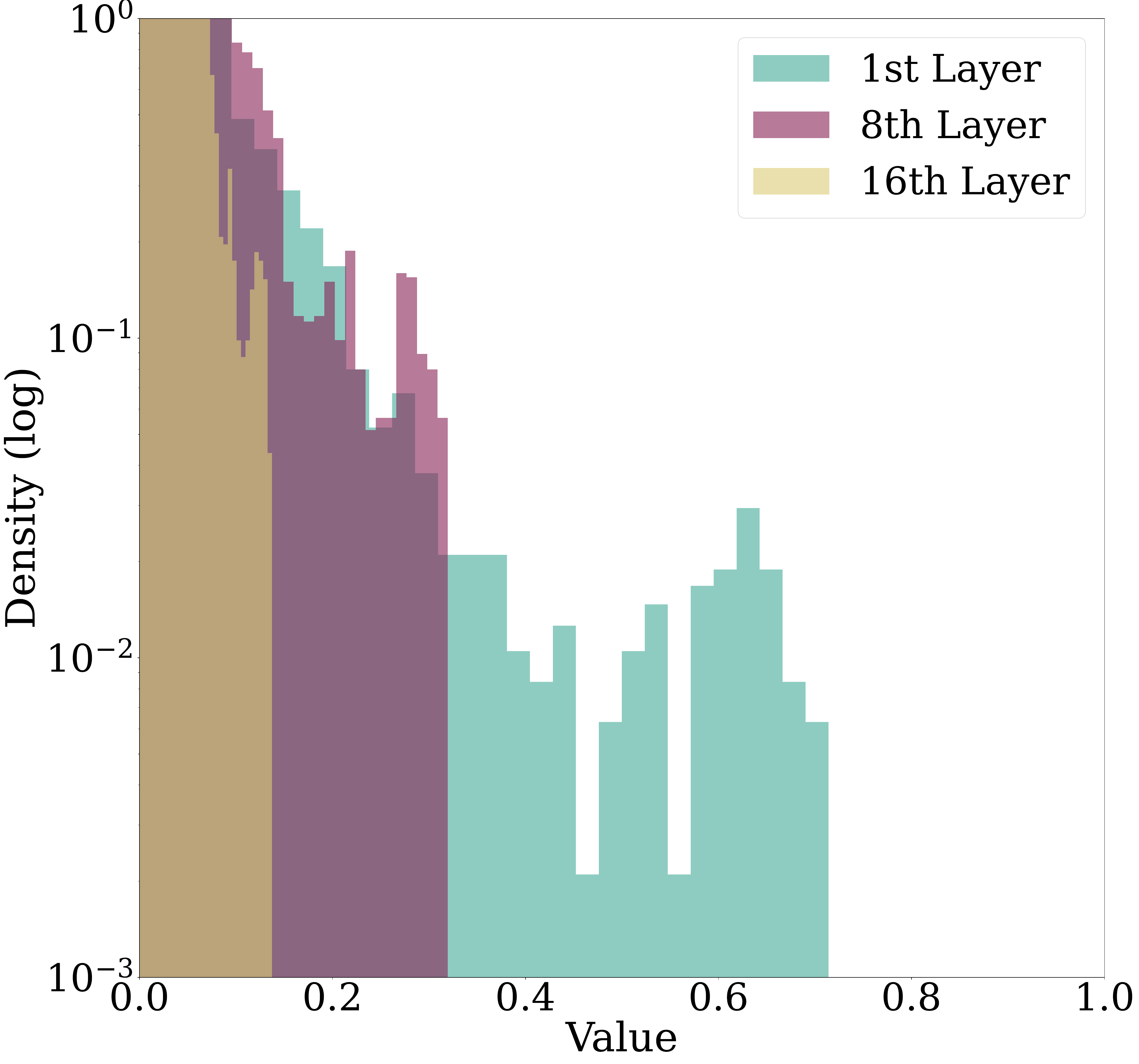}
\caption{Using Multi-Layer LAM}
\end{subfigure}%
\caption{\small{\bf Attention Weight Distribution.} We investigate the impact of our Learnable Attention Mask (LAM) on attention weight distribution during AD generation task. Utilizing LAM, attention weights decrease in value as they traverse deeper layers, with many approaching zero. This enhances efficiency in model training by reducing unnecessary computations. The attention weights were collected by doing forward pass of 64 samples.}
\label{fig:attn}
%\vspace{-0.3cm}
\end{figure}

%% file: figures/6_appendix_fig1.tex
\begin{figure}[ht]
    \centering
    \begin{lstlisting}[language=bash, breaklines=true]
    Below is an instruction that describes a task
    ### Instruction:
    Generate caption of this video.
    ### Response:
    \end{lstlisting}
    \caption{\textbf{Prompt for Audio Description Generation} The caption provided outlines the prompt utilized to activate the functionality of Audio Description generation employing the LLaMA model.}
    \label{code:prompt}
\end{figure}

%% file: figures/6_appendeix_qualitative.tex
\begin{figure*}[ht!]
    \centering
     %\vspace{-0.4cm}
    \includegraphics[width=0.99\textwidth]{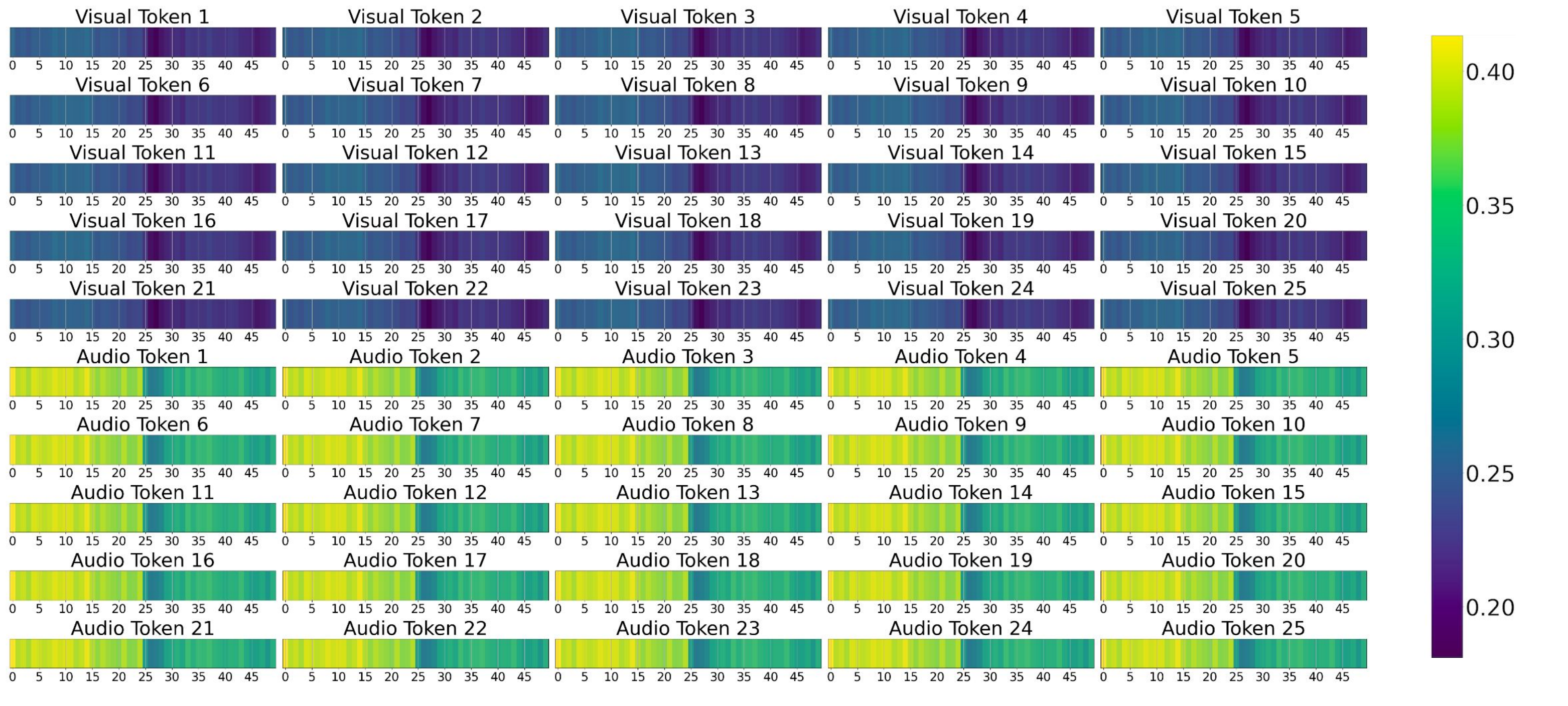}
    % \vspace{-0.3cm}
   \caption{\small\textbf{Analysis of LAM Failure Example in Audio Description Generation.} This plot shows the learnable mask (LAM's output) from the example in Figure~\ref{fig:qualitative_analysis}, with the visual features remaining the same but the audio tokens corresponding to the last 25 samples from the movie's credits that only include soundtrack. While the visual part assigns low values appropriately, the audio part fails to choose appropriate values, assigning mid-range values from the distribution. Note that video tokens are represented on the x-axis from 0 to 24, while audio tokens range from 25 to 50 on the same axis.}
    \label{fig:failure_analysis}
    \vspace{-0.5cm}
\end{figure*}

%% file: neurips_2024.bbl
\begin{thebibliography}{46}
\providecommand{\natexlab}[1]{#1}
\providecommand{\url}[1]{\texttt{#1}}
\expandafter\ifx\csname urlstyle\endcsname\relax
  \providecommand{\doi}[1]{doi: #1}\else
  \providecommand{\doi}{doi: \begingroup \urlstyle{rm}\Url}\fi

\bibitem[Anderson et~al.(2016)Anderson, Fernando, Johnson, and Gould]{spice}
Peter Anderson, Basura Fernando, Mark Johnson, and Stephen Gould.
\newblock {SPICE:} semantic propositional image caption evaluation.
\newblock In Bastian Leibe, Jiri Matas, Nicu Sebe, and Max Welling, editors, \emph{Computer Vision - {ECCV} 2016 - 14th European Conference, Amsterdam, The Netherlands, October 11-14, 2016, Proceedings, Part {V}}, volume 9909 of \emph{Lecture Notes in Computer Science}, pages 382--398. Springer, 2016.
\newblock \doi{10.1007/978-3-319-46454-1\_24}.
\newblock URL \url{https://doi.org/10.1007/978-3-319-46454-1\_24}.

\bibitem[Arandjelovic and Zisserman(2017)]{openl32}
Relja Arandjelovic and Andrew Zisserman.
\newblock Look, listen and learn.
\newblock In \emph{2017 IEEE International Conference on Computer Vision (ICCV)}, pages 609--617, 2017.
\newblock \doi{10.1109/ICCV.2017.73}.

\bibitem[Barrios et~al.(2023)Barrios, Soldan, Ceballos-Arroyo, Heilbron, and Ghanem]{Barrios_2023_ICCV}
Wayner Barrios, Mattia Soldan, Alberto~Mario Ceballos-Arroyo, Fabian~Caba Heilbron, and Bernard Ghanem.
\newblock Localizing moments in long video via multimodal guidance.
\newblock In \emph{Proceedings of the IEEE/CVF International Conference on Computer Vision (ICCV)}, pages 13667--13678, October 2023.

\bibitem[Chen et~al.(2021)Chen, Fan, and Panda]{xattn}
C.~Chen, Q.~Fan, and R.~Panda.
\newblock Crossvit: Cross-attention multi-scale vision transformer for image classification.
\newblock In \emph{2021 IEEE/CVF International Conference on Computer Vision (ICCV)}, pages 347--356, Los Alamitos, CA, USA, oct 2021. IEEE Computer Society.
\newblock \doi{10.1109/ICCV48922.2021.00041}.
\newblock URL \url{https://doi.ieeecomputersociety.org/10.1109/ICCV48922.2021.00041}.

\bibitem[Chen et~al.(2023)Chen, Liu, Hao, Nie, Arap, and Hamid]{Chen_2023_CVPR}
Shixing Chen, Chun-Hao Liu, Xiang Hao, Xiaohan Nie, Maxim Arap, and Raffay Hamid.
\newblock Movies2scenes: Using movie metadata to learn scene representation.
\newblock In \emph{Proceedings of the IEEE/CVF Conference on Computer Vision and Pattern Recognition (CVPR)}, pages 6535--6544, June 2023.

\bibitem[Chen et~al.(2020)Chen, Kornblith, Norouzi, and Hinton]{simclr}
Ting Chen, Simon Kornblith, Mohammad Norouzi, and Geoffrey~E. Hinton.
\newblock A simple framework for contrastive learning of visual representations.
\newblock \emph{ArXiv}, abs/2002.05709, 2020.
\newblock URL \url{https://api.semanticscholar.org/CorpusID:211096730}.

\bibitem[Cramer et~al.(2019)Cramer, Wu, Salamon, and Bello]{openl31}
Jason Cramer, Ho-Hsiang Wu, Justin Salamon, and Juan~Pablo Bello.
\newblock Look, listen, and learn more: Design choices for deep audio embeddings.
\newblock In \emph{ICASSP 2019 - 2019 IEEE International Conference on Acoustics, Speech and Signal Processing (ICASSP)}, pages 3852--3856, 2019.
\newblock \doi{10.1109/ICASSP.2019.8682475}.

\bibitem[Deng et~al.(2009)Deng, Dong, Socher, Li, Li, and Fei-Fei]{imagenet}
Jia Deng, Wei Dong, Richard Socher, Li-Jia Li, K.~Li, and Li~Fei-Fei.
\newblock Imagenet: A large-scale hierarchical image database.
\newblock \emph{2009 IEEE Conference on Computer Vision and Pattern Recognition}, pages 248--255, 2009.
\newblock URL \url{https://api.semanticscholar.org/CorpusID:57246310}.

\bibitem[Fan et~al.(2021)Fan, Gong, Liu, Wei, Wang, Jiao, Duan, Zhang, and Huang]{mask_networks}
Zhihao Fan, Yeyun Gong, Dayiheng Liu, Zhongyu Wei, Siyuan Wang, Jian Jiao, Nan Duan, Ruofei Zhang, and Xuanjing Huang.
\newblock Mask attention networks: Rethinking and strengthen transformer.
\newblock In Kristina Toutanova, Anna Rumshisky, Luke Zettlemoyer, Dilek Hakkani-Tur, Iz~Beltagy, Steven Bethard, Ryan Cotterell, Tanmoy Chakraborty, and Yichao Zhou, editors, \emph{Proceedings of the 2021 Conference of the North American Chapter of the Association for Computational Linguistics: Human Language Technologies}, pages 1692--1701, Online, June 2021. Association for Computational Linguistics.
\newblock \doi{10.18653/v1/2021.naacl-main.135}.
\newblock URL \url{https://aclanthology.org/2021.naacl-main.135}.

\bibitem[Feichtenhofer et~al.(2018)Feichtenhofer, Fan, Malik, and He]{SlowFastNF}
Christoph Feichtenhofer, Haoqi Fan, Jitendra Malik, and Kaiming He.
\newblock Slowfast networks for video recognition.
\newblock \emph{2019 IEEE/CVF International Conference on Computer Vision (ICCV)}, pages 6201--6210, 2018.

\bibitem[Gao et~al.(2023)Gao, Han, Zhang, Lin, Geng, Zhou, Zhang, Lu, He, Yue, Li, and Qiao]{llama_adapterv2}
Peng Gao, Jiaming Han, Renrui Zhang, Ziyi Lin, Shijie Geng, Aojun Zhou, Wei Zhang, Pan Lu, Conghui He, Xiangyu Yue, Hongsheng Li, and Yu~Qiao.
\newblock Llama-adapter v2: Parameter-efficient visual instruction model.
\newblock \emph{arXiv preprint arXiv:2304.15010}, 2023.

\bibitem[Han et~al.(2023{\natexlab{a}})Han, Zhang, Shao, Gao, Xu, Xiao, Zhang, Liu, Wen, Guo, Lu, Ren, Wen, Chen, Yue, Li, and Qiao]{han2023imagebindllm}
Jiaming Han, Renrui Zhang, Wenqi Shao, Peng Gao, Peng Xu, Han Xiao, Kaipeng Zhang, Chris Liu, Song Wen, Ziyu Guo, Xudong Lu, Shuai Ren, Yafei Wen, Xiaoxin Chen, Xiangyu Yue, Hongsheng Li, and Yu~Qiao.
\newblock Imagebind-llm: Multi-modality instruction tuning, 2023{\natexlab{a}}.

\bibitem[Han et~al.(2023{\natexlab{b}})Han, Bain, Nagrani, Varol, Xie, and Zisserman]{autoad1}
Tengda Han, Max Bain, Arsha Nagrani, G\"ul Varol, Weidi Xie, and Andrew Zisserman.
\newblock {AutoAD}: Movie description in context.
\newblock In \emph{CVPR}, 2023{\natexlab{b}}.

\bibitem[Han et~al.(2023{\natexlab{c}})Han, Bain, Nagrani, Varol, Xie, and Zisserman]{autoad2}
Tengda Han, Max Bain, Arsha Nagrani, G\"ul Varol, Weidi Xie, and Andrew Zisserman.
\newblock {AutoAD II: The Sequel} - who, when, and what in movie audio description.
\newblock In \emph{ICCV}, 2023{\natexlab{c}}.

\bibitem[He et~al.(2019)He, Fan, Wu, Xie, and Girshick]{moco}
Kaiming He, Haoqi Fan, Yuxin Wu, Saining Xie, and Ross~B. Girshick.
\newblock Momentum contrast for unsupervised visual representation learning.
\newblock \emph{2020 IEEE/CVF Conference on Computer Vision and Pattern Recognition (CVPR)}, pages 9726--9735, 2019.
\newblock URL \url{https://api.semanticscholar.org/CorpusID:207930212}.

\bibitem[He et~al.(2021)He, Chen, Xie, Li, Doll'ar, and Girshick]{mae}
Kaiming He, Xinlei Chen, Saining Xie, Yanghao Li, Piotr Doll'ar, and Ross~B. Girshick.
\newblock Masked autoencoders are scalable vision learners.
\newblock \emph{2022 IEEE/CVF Conference on Computer Vision and Pattern Recognition (CVPR)}, pages 15979--15988, 2021.

\bibitem[Hu et~al.(2021)Hu, Shen, Wallis, Allen-Zhu, Li, Wang, and Chen]{lora}
J.~Edward Hu, Yelong Shen, Phillip Wallis, Zeyuan Allen-Zhu, Yuanzhi Li, Shean Wang, and Weizhu Chen.
\newblock Lora: Low-rank adaptation of large language models.
\newblock \emph{ArXiv}, abs/2106.09685, 2021.
\newblock URL \url{https://api.semanticscholar.org/CorpusID:235458009}.

\bibitem[Hu et~al.(2023)Hu, Lan, Wang, Xu, Lim, Lee, Bing, and Poria]{hu2023llm}
Zhiqiang Hu, Yihuai Lan, Lei Wang, Wanyu Xu, Ee-Peng Lim, Roy Ka-Wei Lee, Lidong Bing, and Soujanya Poria.
\newblock Llm-adapters: An adapter family for parameter-efficient fine-tuning of large language models.
\newblock \emph{arXiv preprint arXiv:2304.01933}, 2023.

\bibitem[Islam et~al.(2023)Islam, Hasan, Athrey, Braskich, and Bertasius]{Islam_2023_CVPR}
Md~Mohaiminul Islam, Mahmudul Hasan, Kishan~Shamsundar Athrey, Tony Braskich, and Gedas Bertasius.
\newblock Efficient movie scene detection using state-space transformers.
\newblock In \emph{Proceedings of the IEEE/CVF Conference on Computer Vision and Pattern Recognition (CVPR)}, pages 18749--18758, June 2023.

\bibitem[Kamath et~al.(2021)Kamath, Singh, LeCun, Misra, Synnaeve, and Carion]{Kamath2021MDETRM}
Aishwarya Kamath, Mannat Singh, Yann LeCun, Ishan Misra, Gabriel Synnaeve, and Nicolas Carion.
\newblock Mdetr - modulated detection for end-to-end multi-modal understanding.
\newblock \emph{2021 IEEE/CVF International Conference on Computer Vision (ICCV)}, pages 1760--1770, 2021.

\bibitem[Khosla et~al.(2020)Khosla, Teterwak, Wang, Sarna, Tian, Isola, Maschinot, Liu, and Krishnan]{khosla2020supervised}
Prannay Khosla, Piotr Teterwak, Chen Wang, Aaron Sarna, Yonglong Tian, Phillip Isola, Aaron Maschinot, Ce~Liu, and Dilip Krishnan.
\newblock Supervised contrastive learning.
\newblock \emph{arXiv preprint arXiv:2004.11362}, 2020.

\bibitem[Lavie and Agarwal(2007)]{meteor}
Alon Lavie and Abhaya Agarwal.
\newblock Meteor: an automatic metric for mt evaluation with high levels of correlation with human judgments.
\newblock In \emph{Proceedings of the Second Workshop on Statistical Machine Translation}, StatMT '07, page 228–231, USA, 2007. Association for Computational Linguistics.

\bibitem[Lee et~al.(2021)Lee, Jain, Park, and Yun]{Lee2021CrossAttentionalAF}
Jun-Tae Lee, Mihir Jain, Hyoungwoo Park, and Sungrack Yun.
\newblock Cross-attentional audio-visual fusion for weakly-supervised action localization.
\newblock In \emph{International Conference on Learning Representations}, 2021.
\newblock URL \url{https://api.semanticscholar.org/CorpusID:235614339}.

\bibitem[Lei et~al.(2021)Lei, Berg, and Bansal]{moment-detr}
Jie Lei, Tamara~L Berg, and Mohit Bansal.
\newblock Detecting moments and highlights in videos via natural language queries.
\newblock In M.~Ranzato, A.~Beygelzimer, Y.~Dauphin, P.S. Liang, and J.~Wortman Vaughan, editors, \emph{Advances in Neural Information Processing Systems}, volume~34, pages 11846--11858. Curran Associates, Inc., 2021.
\newblock URL \url{https://proceedings.neurips.cc/paper/2021/file/62e0973455fd26eb03e91d5741a4a3bb-Paper.pdf}.

\bibitem[Li et~al.(2021)Li, Chen, Yang, Li, Zhu, Zhao, Deng, Wu, Zhao, Tang, and Wang]{MST}
Zhaowen Li, Zhiyang Chen, Fan Yang, Wei Li, Yousong Zhu, Chaoyang Zhao, Rui Deng, Liwei Wu, Rui Zhao, Ming Tang, and Jinqiao Wang.
\newblock Mst: Masked self-supervised transformer for visual representation.
\newblock In M.~Ranzato, A.~Beygelzimer, Y.~Dauphin, P.S. Liang, and J.~Wortman Vaughan, editors, \emph{Advances in Neural Information Processing Systems}, volume~34, pages 13165--13176. Curran Associates, Inc., 2021.
\newblock URL \url{https://proceedings.neurips.cc/paper_files/paper/2021/file/6dbbe6abe5f14af882ff977fc3f35501-Paper.pdf}.

\bibitem[Lin(2004)]{rouge}
Chin-Yew Lin.
\newblock {ROUGE}: A package for automatic evaluation of summaries.
\newblock In \emph{Text Summarization Branches Out}, pages 74--81, Barcelona, Spain, July 2004. Association for Computational Linguistics.
\newblock URL \url{https://aclanthology.org/W04-1013}.

\bibitem[Lin et~al.(2022)Lin, Li, Lin, Ahmed, Gan, Liu, Lu, and Wang]{swinbert}
Kevin Lin, Linjie Li, Chung-Ching Lin, Faisal Ahmed, Zhe Gan, Zicheng Liu, Yumao Lu, and Lijuan Wang.
\newblock Swinbert: End-to-end transformers with sparse attention for video captioning.
\newblock In \emph{CVPR}, 2022.

\bibitem[Lin and Joe(2023)]{adapt_mask}
Te~Lin and Inwhee Joe.
\newblock An adaptive masked attention mechanism to act on the local text in a global context for aspect-based sentiment analysis.
\newblock \emph{IEEE Access}, 11:\penalty0 43055--43066, 2023.
\newblock \doi{10.1109/ACCESS.2023.3270927}.

\bibitem[Luo et~al.(2021)Luo, Ji, Zhong, Chen, Lei, Duan, and Li]{luo2021clip4clip}
Huaishao Luo, Lei Ji, Ming Zhong, Yang Chen, Wen Lei, Nan Duan, and Tianrui Li.
\newblock Clip4clip: An empirical study of clip for end to end video clip retrieval.
\newblock \emph{arXiv preprint arXiv:2104.08860}, 2021.

\bibitem[Moon et~al.(2023)Moon, Hyun, Park, Park, and Heo]{qd_detr}
WonJun Moon, Sangeek Hyun, SangUk Park, Dongchan Park, and Jae-Pil Heo.
\newblock Query-dependent video representation for moment retrieval and highlight detection.
\newblock In \emph{Proceedings of the IEEE/CVF Conference on Computer Vision and Pattern Recognition}, pages 23023--23033, 2023.

\bibitem[Papineni et~al.(2002)Papineni, Roukos, Ward, and Zhu]{bleu}
Kishore Papineni, Salim Roukos, Todd Ward, and Wei-Jing Zhu.
\newblock Bleu: a method for automatic evaluation of machine translation.
\newblock In \emph{Proceedings of the 40th Annual Meeting on Association for Computational Linguistics}, ACL '02, page 311–318, USA, 2002. Association for Computational Linguistics.
\newblock \doi{10.3115/1073083.1073135}.
\newblock URL \url{https://doi.org/10.3115/1073083.1073135}.

\bibitem[Radford et~al.(2021)Radford, Kim, Hallacy, Ramesh, Goh, Agarwal, Sastry, Askell, Mishkin, Clark, Krueger, and Sutskever]{clip}
Alec Radford, Jong~Wook Kim, Chris Hallacy, Aditya Ramesh, Gabriel Goh, Sandhini Agarwal, Girish Sastry, Amanda Askell, Pamela Mishkin, Jack Clark, Gretchen Krueger, and Ilya Sutskever.
\newblock Learning transferable visual models from natural language supervision.
\newblock In \emph{International Conference on Machine Learning}, 2021.
\newblock URL \url{https://api.semanticscholar.org/CorpusID:231591445}.

\bibitem[Rende et~al.(2024)Rende, Gerace, Laio, and Goldt]{self_mask}
Riccardo Rende, Federica Gerace, Alessandro Laio, and Sebastian Goldt.
\newblock What does self-attention learn from masked language modelling?, 2024.

\bibitem[Rohrbach et~al.(2015)Rohrbach, Rohrbach, Tandon, and Schiele]{lsmdc}
Anna Rohrbach, Marcus Rohrbach, Niket Tandon, and Bernt Schiele.
\newblock A dataset for movie description.
\newblock \emph{2015 IEEE Conference on Computer Vision and Pattern Recognition (CVPR)}, pages 3202--3212, 2015.
\newblock URL \url{https://api.semanticscholar.org/CorpusID:15184723}.

\bibitem[Soldan et~al.(2022)Soldan, Pardo, Alc\'azar, Caba, Zhao, Giancola, and Ghanem]{mad}
Mattia Soldan, Alejandro Pardo, Juan~Le\'on Alc\'azar, Fabian Caba, Chen Zhao, Silvio Giancola, and Bernard Ghanem.
\newblock Mad: A scalable dataset for language grounding in videos from movie audio descriptions.
\newblock In \emph{Proceedings of the IEEE/CVF Conference on Computer Vision and Pattern Recognition (CVPR)}, pages 5026--5035, June 2022.

\bibitem[Tang et~al.(2021)Tang, Wu, Zhang, Zhang, and Jiang]{MDMAN}
Jingfan Tang, Xinqiang Wu, Min Zhang, Xiujie Zhang, and Ming Jiang.
\newblock Multiway dynamic mask attention networks for natural language inference.
\newblock \emph{J. Comp. Methods in Sci. and Eng.}, 21\penalty0 (1):\penalty0 151–162, jan 2021.
\newblock ISSN 1472-7978.
\newblock \doi{10.3233/JCM-204451}.
\newblock URL \url{https://doi.org/10.3233/JCM-204451}.

\bibitem[Touvron et~al.(2023)Touvron, Lavril, Izacard, Martinet, Lachaux, Lacroix, Rozi{\`e}re, Goyal, Hambro, Azhar, Rodriguez, Joulin, Grave, and Lample]{llama}
Hugo Touvron, Thibaut Lavril, Gautier Izacard, Xavier Martinet, Marie-Anne Lachaux, Timoth{\'e}e Lacroix, Baptiste Rozi{\`e}re, Naman Goyal, Eric Hambro, Faisal Azhar, Aurelien Rodriguez, Armand Joulin, Edouard Grave, and Guillaume Lample.
\newblock Llama: Open and efficient foundation language models.
\newblock \emph{ArXiv}, abs/2302.13971, 2023.
\newblock URL \url{https://api.semanticscholar.org/CorpusID:257219404}.

\bibitem[Vaswani et~al.(2017)Vaswani, Shazeer, Parmar, Uszkoreit, Jones, Gomez, Kaiser, and Polosukhin]{transformer}
Ashish Vaswani, Noam Shazeer, Niki Parmar, Jakob Uszkoreit, Llion Jones, Aidan~N Gomez, \L~ukasz Kaiser, and Illia Polosukhin.
\newblock Attention is all you need.
\newblock In I.~Guyon, U.~Von Luxburg, S.~Bengio, H.~Wallach, R.~Fergus, S.~Vishwanathan, and R.~Garnett, editors, \emph{Advances in Neural Information Processing Systems}, volume~30. Curran Associates, Inc., 2017.
\newblock URL \url{https://proceedings.neurips.cc/paper_files/paper/2017/file/3f5ee243547dee91fbd053c1c4a845aa-Paper.pdf}.

\bibitem[Vedantam et~al.(2014)Vedantam, Zitnick, and Parikh]{cider}
Ramakrishna Vedantam, C.~Lawrence Zitnick, and Devi Parikh.
\newblock Cider: Consensus-based image description evaluation.
\newblock \emph{2015 IEEE Conference on Computer Vision and Pattern Recognition (CVPR)}, pages 4566--4575, 2014.
\newblock URL \url{https://api.semanticscholar.org/CorpusID:9026666}.

\bibitem[Wei et~al.(2020)Wei, Zhang, Li, Zhang, and Wu]{multixatt}
Xi~Wei, Tianzhu Zhang, Yan Li, Yongdong Zhang, and Feng Wu.
\newblock Multi-modality cross attention network for image and sentence matching.
\newblock In \emph{2020 IEEE/CVF Conference on Computer Vision and Pattern Recognition (CVPR)}, pages 10938--10947, 2020.
\newblock \doi{10.1109/CVPR42600.2020.01095}.

\bibitem[Xiao et~al.(2022)Xiao, Kundu, Tighe, and Modolo]{Xiao_2022_CVPR}
Fanyi Xiao, Kaustav Kundu, Joseph Tighe, and Davide Modolo.
\newblock Hierarchical self-supervised representation learning for movie understanding.
\newblock In \emph{Proceedings of the IEEE/CVF Conference on Computer Vision and Pattern Recognition (CVPR)}, pages 9727--9736, June 2022.

\bibitem[Xu et~al.(2016)Xu, Mei, Yao, and Rui]{msrvtt}
Jun Xu, Tao Mei, Ting Yao, and Yong Rui.
\newblock Msr-vtt: A large video description dataset for bridging video and language.
\newblock \emph{2016 IEEE Conference on Computer Vision and Pattern Recognition (CVPR)}, pages 5288--5296, 2016.

\bibitem[Yi-Lin~Sung(2022)]{sung2022vladapter}
Mohit~Bansal Yi-Lin~Sung, Jaemin~Cho.
\newblock Vl-adapter: Parameter-efficient transfer learning for vision-and-language tasks.
\newblock In \emph{CVPR}, 2022.

\bibitem[Zhang et~al.(2023{\natexlab{a}})Zhang, Li, and Bing]{zhang2023videollama}
Hang Zhang, Xin Li, and Lidong Bing.
\newblock Video-llama: An instruction-tuned audio-visual language model for video understanding, 2023{\natexlab{a}}.

\bibitem[Zhang et~al.(2023{\natexlab{b}})Zhang, Han, Liu, Gao, Zhou, Hu, Yan, Lu, Li, and Qiao]{llama_adapterv1}
Renrui Zhang, Jiaming Han, Chris Liu, Peng Gao, Aojun Zhou, Xiangfei Hu, Shilin Yan, Pan Lu, Hongsheng Li, and Yu~Qiao.
\newblock Llama-adapter: Efficient fine-tuning of language models with zero-init attention.
\newblock \emph{arXiv preprint arXiv:2303.16199}, 2023{\natexlab{b}}.

\bibitem[Zhang et~al.(2020)Zhang, Kishore, Wu, Weinberger, and Artzi]{bertscore}
Tianyi Zhang, Varsha Kishore, Felix Wu, Kilian~Q. Weinberger, and Yoav Artzi.
\newblock Bertscore: Evaluating text generation with {BERT}.
\newblock In \emph{8th International Conference on Learning Representations, {ICLR} 2020, Addis Ababa, Ethiopia, April 26-30, 2020}. OpenReview.net, 2020.
\newblock URL \url{https://openreview.net/forum?id=SkeHuCVFDr}.

\end{thebibliography}
